\documentclass[11pt]{article}

\usepackage[final]{acl}

\usepackage{times}
\usepackage{latexsym}
\usepackage[T1]{fontenc}
\usepackage[utf8]{inputenc}
\usepackage{microtype}
\usepackage{inconsolata}
\usepackage{graphicx}

\usepackage{amsmath}
\usepackage{amssymb}   %
\usepackage{enumitem}

\usepackage{xcolor}
\usepackage{pifont}
\usepackage{makecell}
\usepackage{cleveref}
\usepackage[most]{tcolorbox}
\usepackage{mdframed}
\usepackage{colortbl}

\definecolor{darkgreen}{RGB}{0,100,0}
\definecolor{darkblue}{RGB}{0,0,139}%
\definecolor{navyblue}{RGB}{0,0,128}%

\newcommand{\cmark}{\ding{51}}
\newcommand{\xmark}{\ding{55}}
\newcommand{\formatcorrectline}[1]{\multicolumn{1}{c}{\small\textit{#1}}}
\newcommand{\formatcorrecttext}{\formatcorrectline{Correct Answers}}
\newcommand{\formatincorrecttext}{\formatcorrectline{Incorrect Answers}}

\title{Rethinking Math Reasoning Evaluation: A Robust LLM-as-a-Judge Framework Beyond Symbolic Rigidity}

\author{
Erez Yosef$^{1}$ \quad
Oron Anschel$^{2}$ \quad
Shunit Haviv Hakimi$^{2}$ \quad
Asaf Gendler$^{2}$ \\
\textbf{Adam Botach$^{2}$ \quad
Nimrod Berman$^{3}$ \quad
Igor Kviatkovsky$^{2}$} \\[0.5em]
$^{1}$Tel-Aviv University \quad
$^{2}$Amazon Prime Video \quad
$^{3}$Ben-Gurion University \\[0.5em]
\footnotesize{\texttt{erez.yo@gmail.com \quad
\{oronans,havivs,gendlasa,kabotach,kviat\}@amazon.com \quad
bermann@post.bgu.ac.il}}
}

\begin{document}
\maketitle

\begin{abstract}
Recent advancements in large language models have led to significant improvements across various tasks, including mathematical reasoning, which is used to assess models' intelligence in logical reasoning and problem-solving. Models are evaluated on mathematical reasoning benchmarks by verifying the correctness of the final answer against a ground truth answer. A common approach for this verification is based on symbolic mathematics comparison, which fails to generalize across diverse mathematical representations and solution formats.
In this work, we offer a robust and flexible alternative to rule-based symbolic mathematics comparison. We propose an LLM-based evaluation framework for evaluating model-generated answers, enabling accurate evaluation across diverse mathematical representations and answer formats. We present failure cases of symbolic evaluation in two popular frameworks, Lighteval and SimpleRL, and compare them to our approach, demonstrating clear improvements over commonly used methods. Our framework enables more reliable evaluation and benchmarking, leading to more accurate performance monitoring, which is important for advancing mathematical problem-solving and intelligent systems.
\end{abstract}

\section{Introduction}
\label{sec:introduction}
Large language models (LLMs) have demonstrated remarkable progress in recent years, achieving impressive results across a wide range of natural language processing and reasoning tasks.
Among these capabilities, mathematical reasoning is one of the most fundamental and challenging tasks for LLMs and is used to assess models' intelligence and ability to perform complex reasoning tasks.
Recent advances and techniques that have improved LLMs in mathematical reasoning include pre-training on large mathematical corpora \cite{guo2025deepseek, yang2024qwen2}, chain-of-thought prompting which encourages the model to produce step-by-step reasoning of the mathematical solution \cite{wei2022chain}, and fine-tuning with Reinforcement Learning with Verifiable Rewards (RLVR), including methods such as Group Reinforcement Policy Optimization (GRPO) \cite{guo2025deepseek}.

\begin{figure}[t]
\centering
\includegraphics[width=1.0\columnwidth]{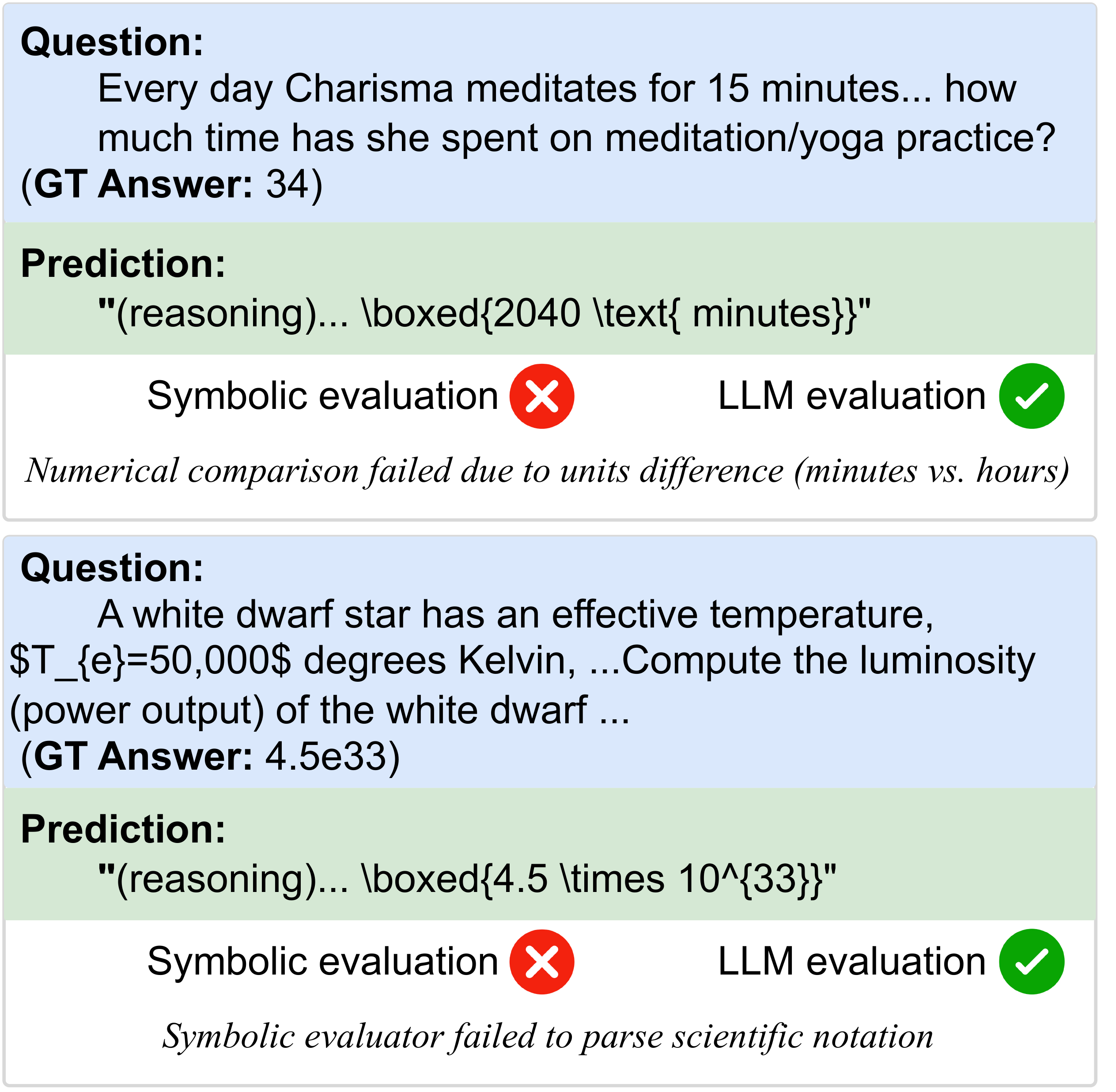}
\caption{Our LLM evaluation approach provides a more robust evaluation compared to traditional symbolic evaluation methods by handling diverse mathematical representations and answer formats. These examples demonstrate the contribution of our approach by correctly evaluating these model predictions for mathematical questions, while the existing numerical comparison approach fails.}
\label{fig:teaser}
\end{figure}

Despite this progress, the evaluation of mathematical reasoning in LLMs remains challenging. Most existing benchmark evaluation methods rely on symbolic mathematics verification or programmatic verification of answers against a reference (ground truth, GT) solution (as Lighteval~\cite{lighteval} and SimpleRL~\cite{zeng2025simplerl}). In this approach, the language model-generated response is parsed to extract the final answer, which is then evaluated as a mathematical expression by symbolic mathematical tools (such as SymPy \cite{sympy}). While effective for well-structured outputs, these methods often fail to generalize to diverse mathematical representations, formats, and approximations generated by the language model. Consequently, models may be inaccurately evaluated due to their answer format, even when the answer is mathematically correct. Most commonly, the models are under-evaluated since the correct answer is predicted in a different format, representation, or units than expected, which causes the symbolic comparison to fail (see \Cref{fig:teaser} and \Cref{tab:examples1}).
Hence, symbolic verification systems are fragile by nature, assuming specific notations and formatting styles as the GT answer. The lack of accurate and robust evaluation introduces significant uncertainty in assessing the genuine mathematical answering correctness of the language models.

Moreover, such mathematical evaluation is also used by the RLVR fine-tuning training approach, which validates the mathematical correctness of the generated outputs of the model to compute a reward for each response. This emphasizes the importance of an accurate response verification method also for the training phase.

In this work, we introduce an LLM-based evaluation framework for mathematical reasoning, in which an LLM acts as an evaluator rather than relying on a pre-defined symbolic verification process. Using the LLM-as-a-judge approach, we leverage the generalization capability and prior knowledge of trained LLMs to evaluate model responses and improve verification accuracy for various response representations, different levels of accuracy, different units, etc.
As an example of different representations of responses and units notation, for a mathematical question with a GT answer of $1000$, some of the correct predictions are:

{\small \begin{verbatim}
"1000",
"1000 \\, \\text{rad/s}",
"\\omega_{ss}=1000 \\mathrm{~r} / \\mathrm{s}",
"1000 \\mathrm{~r} / \\mathrm{s}",
\end{verbatim}
}

\noindent and all these predictions should be evaluated as correct. We present in the following examples more questions and responses where our approach generalizes better to evaluate the model responses compared to the current symbolic approach. To evaluate our method's accuracy and correctness, we conduct meta-evaluations on a small dataset. Since we cannot get a finite set of all possible equivalent representations of a GT answer, having a closed dataset for naive evaluation is not feasible. Hence, for meta-evaluation of our approach, we manually classify a set of responses of the Qwen2.5-7B model for a small set of questions. By this approach, we achieve a closed set of labeled representations that enable us to perform a numerical evaluation. We show a significant improvement over the current method and an ablation study that validates our pipeline design approach.

Our proposed framework addresses these challenges through an LLM-based evaluation approach. LLM-as-a-judge methods are known to exhibit several limitations, including confirmation bias toward ground-truth annotations, positional bias favoring responses based on their ordering, and non-deterministic judgments. To mitigate these issues, we designed our pipeline with the following considerations. First, to overcome dataset problems and reduce confirmation bias toward potentially incorrect dataset annotations, we employ independent question answering, where the judge generates its own solution without access to the GT answer. Second, we validate the generated answer against the dataset GT, synthesizing a verified reference answer that accounts for both sources. The validated answers are used for the LLM-as-a-judge predictions evaluation using multiple assessments of each prediction with majority voting, which improves robustness and reduces variance in judgments. To address positional bias, a known limitation where LLM judges favor responses based on their position in the prompt, we employ random sampling and shuffling of responses across evaluation calls. This design enables flexible evaluation across diverse mathematical representations, notation conventions, unit variations, and precision levels, while maintaining high accuracy through the semantic understanding capabilities of LLMs rather than brittle symbolic matching rules.

Due to the generative nature of language models and their inherent randomness in response generation, models produce different outputs for the same input across multiple sampling attempts. Hence, we evaluate model performance across multiple generated responses by employing the pass@k metric \cite{Chen2021EvaluatingLL}, which measures the probability that at least one correct solution appears among k generated samples. This metric is particularly valuable for assessing the diversity and reliability of model outputs in mathematical reasoning tasks.

\begin{table*}[t]
\centering
\caption{Evaluation examples showing the proposed LLM-as-a-judge evaluation against existing symbolic evaluation. Our approach presents improved accuracy and robustness to format differences over the baseline evaluation.}
\setlength{\tabcolsep}{4pt}
\label{tab:examples1}
\begin{tabular}{cccc}
\hline
GT Answer & Prediction & Symbolic & LLM \small (Ours) \\
\hline
\textcolor{navyblue}{np.arcsin(10/13)} & \small\verb|\arcsin{\frac{1}{1.3}}| & \Large \textcolor{red}{\xmark} & \Large \textcolor{darkgreen}{\cmark} \\
\multicolumn{4}{l}{\footnotesize \textit{Symbolic evaluator cannot parse NumPy function syntax in GT answer}} \\
\hline
\textcolor{navyblue}{$I(0)e^{-t/RC}$} & \small\verb|I(t) = I_0 e^{-\frac{t}{RC}}| & \Large \textcolor{red}{\xmark} & \Large \textcolor{darkgreen}{\cmark} \\
\multicolumn{4}{l}{\footnotesize \textit{Symbolic evaluator fails to verify prediction with equation formatting}} \\
\hline
\textcolor{navyblue}{$x/2-1/4+ce^{-2x}$} & \small\verb|y = \frac{1}{2} x - \frac{1}{4} + Ce^{-2x}| & \Large \textcolor{red}{\xmark} & \Large \textcolor{darkgreen}{\cmark} \\
\multicolumn{4}{l}{\footnotesize \textit{Symbolic evaluator fails with different variable names and formatting}} \\
\hline
\textcolor{navyblue}{$x^{\prime}+\frac{r}{V}x-rc=0$} & \small\verb|\frac{dx}{dt} + \frac{r}{V}x = rc| & \Large \textcolor{red}{\xmark} & \Large \textcolor{darkgreen}{\cmark} \\
\multicolumn{4}{l}{\footnotesize \textit{Symbolic evaluator fails with different derivative notation}} \\
\hline
\textcolor{navyblue}{0.53} & \small\verb|0.533| & \Large \textcolor{red}{\xmark} & \Large \textcolor{darkgreen}{\cmark} \\
\multicolumn{4}{l}{\footnotesize \textit{Symbolic evaluator fails with precision differences (solution of $\ln(2)/1.3$ up to decimal accuracy)}} \\
\hline
\textcolor{navyblue}{1128} & \small\verb|18 \text{ hours } 48 \text{ minutes}| & \Large \textcolor{red}{\xmark} & \Large \textcolor{darkgreen}{\cmark} \\
\multicolumn{4}{l}{\footnotesize \textit{Question asks "...How much time does..."; symbolic evaluator fails with representation differences (minutes vs textual)}} \\
\hline
\end{tabular}
\end{table*}

\noindent Our main contributions are as follows:
\begin{enumerate}[label=\arabic*., nosep, leftmargin=*]
\item A robust evaluation framework that employs an LLM for answer verification of mathematical questions, removing the dependence on symbolic verification and its limitations.
\item A comprehensive analysis of LLM evaluation reliability across diverse mathematical datasets and models.
\item Empirical results demonstrating improved evaluation accuracy over symbolic methods, supported by diverse failure cases of existing approaches.
\item We introduce a meta-evaluation, showing significant gains over prior methods and validating our pipeline design via ablations.
\end{enumerate}

\section{Related Work}
\label{sec:related}

Our work lies at the intersection of three active research areas: mathematical reasoning in large language models, evaluation methods for mathematical problem-solving, and the emerging paradigm of LLM-as-a-judge for model evaluation.

\noindent\textbf{Mathematical Reasoning.}
Mathematical reasoning has emerged as a fundamental benchmark for evaluating the capabilities of large language models, as it requires precise logical reasoning, multi-step problem solving, and symbolic manipulation~\cite{hendrycks2021measuring, cobbe2021training}. The GSM8K dataset~\cite{cobbe2021training} introduced grade-school level word problems requiring multi-step arithmetic reasoning, while the MATH dataset~\cite{hendrycks2021measuring} extended this to competition-level problems spanning algebra, geometry, and calculus. More recently, Olympiad-level benchmarks such as OlympiadBench~\cite{he2024olympiadbench} and Omni-MATH~\cite{gao2024omni} have been introduced.

Chain-of-thought (CoT) prompting~\cite{wei2022chain} represented a breakthrough in mathematical reasoning, enabling models to decompose complex problems into intermediate reasoning steps and significantly improving accuracy on multi-step problems. Self-consistency~\cite{wang2022self} further enhanced this approach by sampling multiple reasoning paths and selecting the most frequent answer.

Specialized mathematical models have been developed through pre-training on large mathematical corpora. Minerva~\cite{lewkowycz2022solving} demonstrated strong performance by training on scientific and mathematical text. DeepSeekMath~\cite{shao2024deepseekmath} and Qwen2.5-Math~\cite{yang2024qwen2} pushed the boundaries further with domain-specific pre-training and instruction tuning for mathematical reasoning.
More recently, reinforcement learning approaches have shown substantial improvements. DeepSeek-R1~\cite{guo2025deepseek} demonstrates that reinforcement learning can incentivize reasoning capabilities in LLMs without extensive supervised fine-tuning. The GRPO algorithm and RLVR approaches, as implemented in frameworks like SimpleRL~\cite{zeng2025simplerl} and Tulu 3~\cite{lambert2024tulu}, leverage mathematical verification as reward signals for training. These methods critically depend on accurate answer verification, highlighting the importance of robust evaluation methodologies addressed in our work.

\noindent\textbf{Mathematical Answers Evaluation.}
The standard approach for evaluating mathematical reasoning relies on comparing model-generated answers against GT solutions. Most evaluation frameworks, including Lighteval~\cite{lighteval}, SimpleRL~\cite{zeng2025simplerl}, and mathematical benchmarks, employ symbolic mathematics tools such as SymPy~\cite{sympy} for this comparison. While effective for well-structured outputs, these symbolic verification methods face significant limitations when handling diverse mathematical representations, approximate answers, and varying notation conventions~\cite{li2024evaluating, xia2025evaluating}.

Several works have highlighted the fragility of current evaluation methodologies. \citet{vendrow2025large} and \citet{gema2025we} demonstrate that benchmark datasets themselves may contain errors or ambiguities, complicating accurate evaluation. GSM-Symbolic~\cite{mirzadeh2024gsm} reveals that LLMs exhibit significant performance variance across different instantiations of the same problem. The MathEval benchmark~\cite{liu2025matheval} addresses some of these concerns by providing comprehensive evaluation across multiple mathematical domains and employing automated pipelines for answer comparison. However, these approaches still fundamentally rely on symbolic matching, which cannot accommodate the full range of mathematically equivalent answer representations.

Beyond final-answer accuracy, recent work has explored evaluating the quality of reasoning processes. ReasonEval~\cite{xia2025evaluating} introduces validity and redundancy metrics to characterize reasoning quality. Process Reward Models (PRMs) have emerged as a powerful approach for step-level supervision. \citet{lightman2023let} demonstrated that process supervision significantly outperforms outcome supervision for training reliable reward models. Math-Shepherd~\cite{wang2024math} extended this by automatically constructing process-wise supervision data without human annotation. Our work complements these approaches by focusing on robust final-answer verification, rather than reasoning process quality assessment, which remains essential for both benchmark evaluation and reward computation in RLVR training.

\noindent\textbf{LLM-as-a-Judge.}
The LLM-as-a-judge paradigm \cite{zheng2023judging} has emerged as a scalable alternative to human evaluation for assessing model outputs across various tasks. MT-Bench and Chatbot Arena~\cite{zheng2023judging} established foundational methodologies for using LLMs as evaluators, showing strong alignment with human preferences. This approach leverages the generalization capabilities of LLMs to evaluate responses based on semantic understanding rather than exact matching~\cite{li2024llms, gu2024survey}.
LLM judges have been applied to diverse applications, including data labeling~\cite{tan2024large}, model self-improvement through meta-rewarding~\cite{wu2024meta}, and benchmark construction~\cite{li2024crowdsourced}. Several benchmarks have been proposed to evaluate judge quality, including JudgeBench~\cite{tan2024judgebench} and Arena-Hard~\cite{li2024crowdsourced}. \citet{ye2024justice} identify systematic biases in LLM judges, including positional bias that affects evaluation fairness. These biases are particularly concerning for mathematical evaluation, where objective correctness should be the primary criterion.

The application of LLM-as-a-judge to mathematical reasoning is still limited. \citet{stephan2024calculation} examine LLM judges on mathematical reasoning tasks and find that judgment performance correlates strongly with the model's problem-solving capabilities, with judges tending to favor responses from higher-quality models regardless of correctness. 

Our work is the first to comprehensively address the limitations of symbolic verification and apply final-answer LLM-based verification. We suggest a robust approach to handle diverse mathematical representations, notation variations, and approximation differences that symbolic tools cannot accommodate, while reducing the judge biases through a multi-stage pipeline.

\section{Methods}
\label{sec:methods}

To ensure accurate evaluation of the models' performance, as a prior step, we aim to obtain a valid mathematical answer for each question. Prior works \cite{vendrow2025large,gema2025we} have shown that dataset questions may be unclear or incomplete, and that GT answers can be incorrect, issues we also encountered in our analysis (see \Cref{appendix:calrity_score}). This validation process is done in two stages: independent question answering by a strong LLM judge (\Cref{sec:methods:s1}) and validating the answer against the GT of the dataset (\Cref{sec:methods:s2}), to get valid results and significantly reduce model biases as we observed in our prior experiments, which we discuss in more detail in the following subsections. %
After having validated answers for the benchmark questions, we evaluate the model predictions using LLM-as-a-judge approach (\Cref{sec:methods:s3}).
To verify our method and design choices, we apply meta-evaluation on several different settings and ablations as detailed in \Cref{sec:methods:meta-evaluation} on a small set of labeled model predictions. This enables us to perform numerical evaluation and quantify our contribution compared to the baseline method.

\subsection{Independent Question Answering}
\label{sec:methods:s1}
In the first stage, we ask the LLM judge (stronger than the evaluated models) to generate an answer for each question in the dataset (prompted independently) without providing it with the GT answer of the dataset.
We observed that when the dataset answer is provided to the LLM, it tends to accept the dataset answer as correct even for cases where the answer is inaccurate or incorrect. This behavior is primarily observed in challenging questions that require domain-specific knowledge, as opposed to elementary grade school math problems. Hence, we reduce this bias toward the dataset GT answer using this approach. The LLM generates for each question its candidate answer, which will be validated in the next stage with the dataset answer. Due to unclear questions in some benchmarks, as observed in prior works \cite{vendrow2025large,gema2025we}, we ask the LLMs to also produce a question clarity score in the range of 1-10 to evaluate the question clarity and consistency (\Cref{tab:unclear_questions}). For this stage, we set the LLM temperature to 0 to get the highest fidelity by the model and reduce creativity and randomness in the responses.

\begin{figure*}[t]
\centering
\setlength{\tabcolsep}{2pt}   %

\newcommand{\plotwidth}{0.4\textwidth}
\newcommand{\raisetxt}{1.2cm}
\begin{tabular}{cc}
\small SimpleRL Evaluation & \small Lighteval Evaluation \\
\includegraphics[width=\plotwidth]{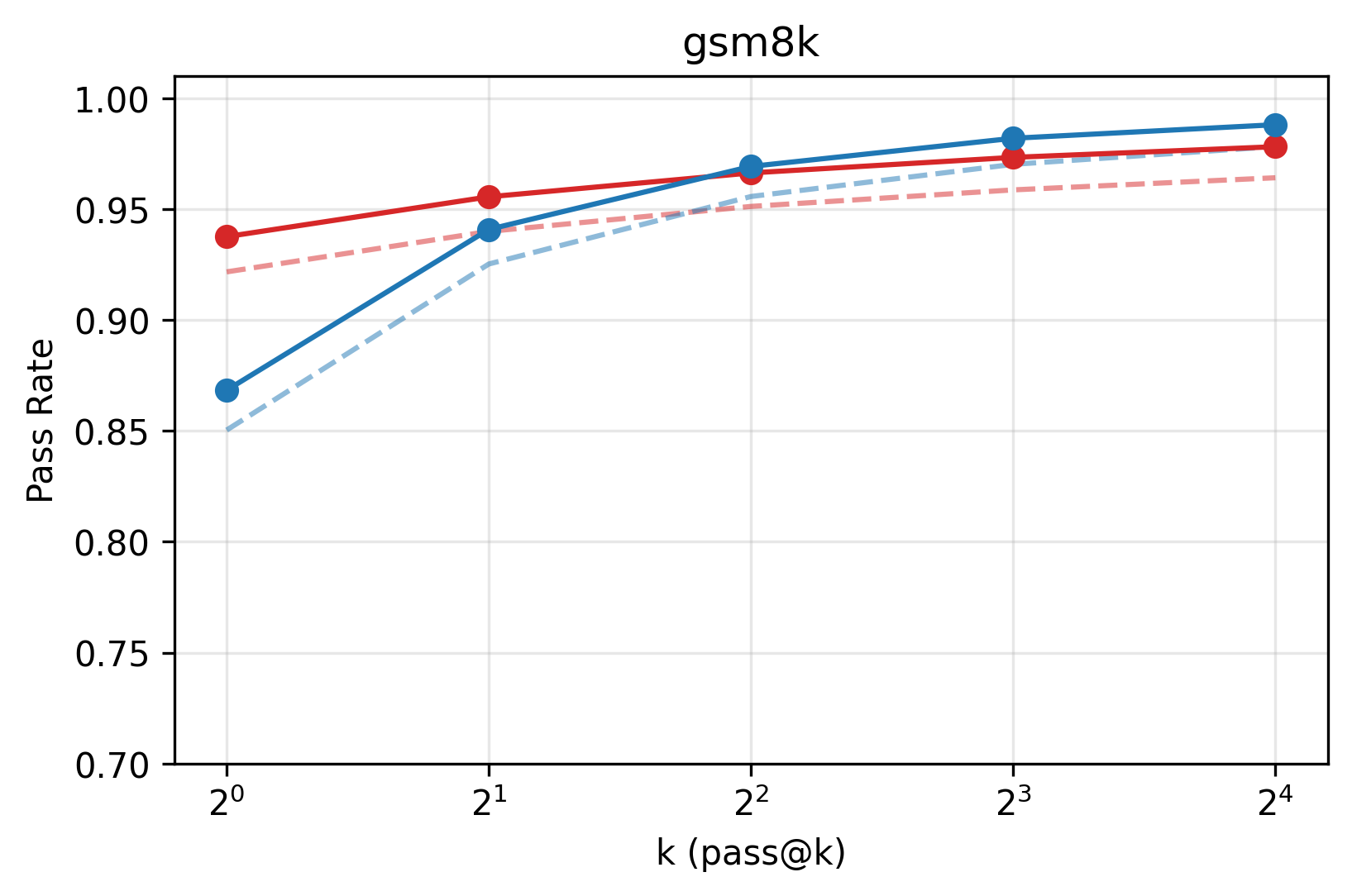} &
\includegraphics[width=\plotwidth]{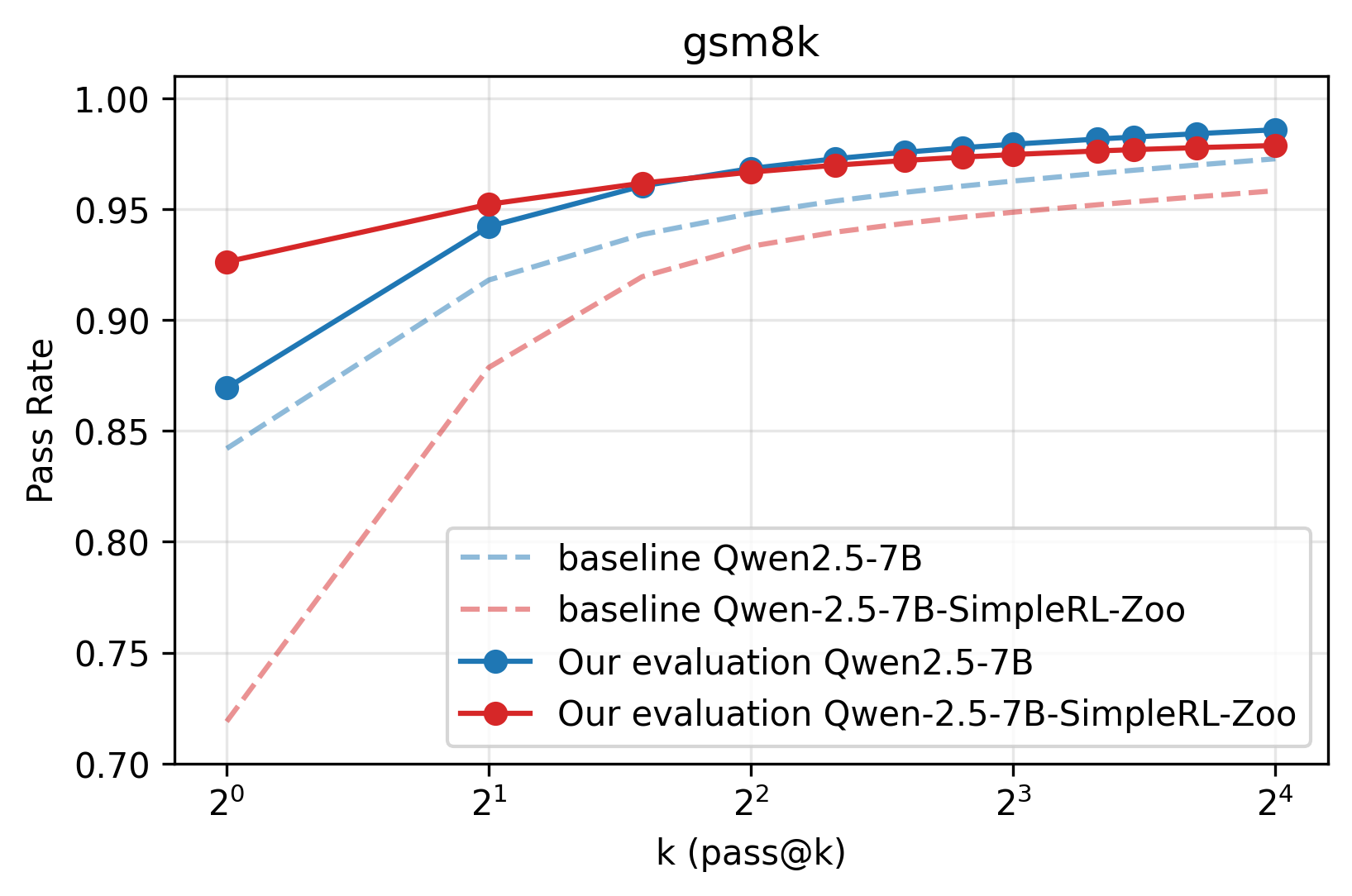} \\
\includegraphics[width=\plotwidth]{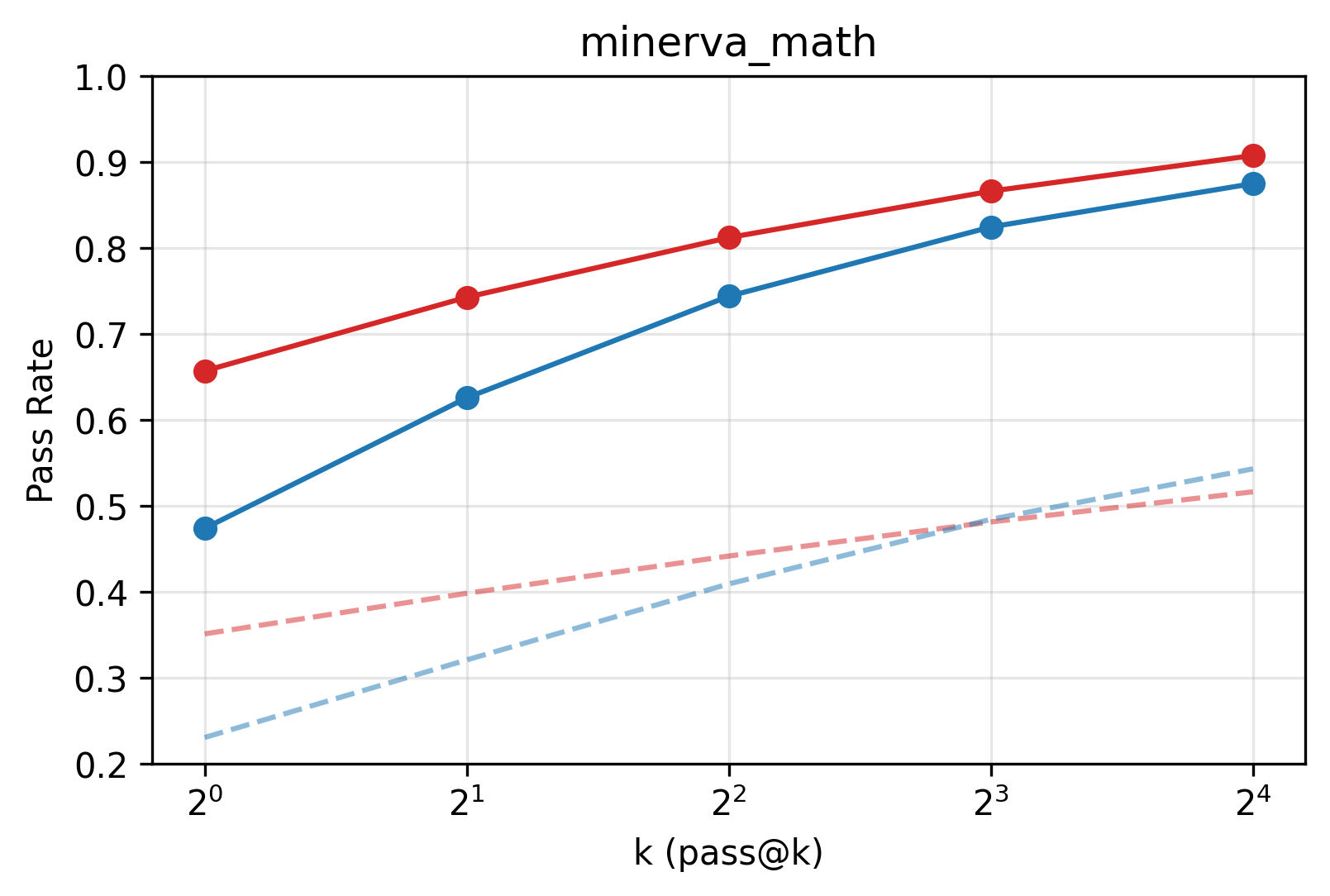} &
\includegraphics[width=\plotwidth]{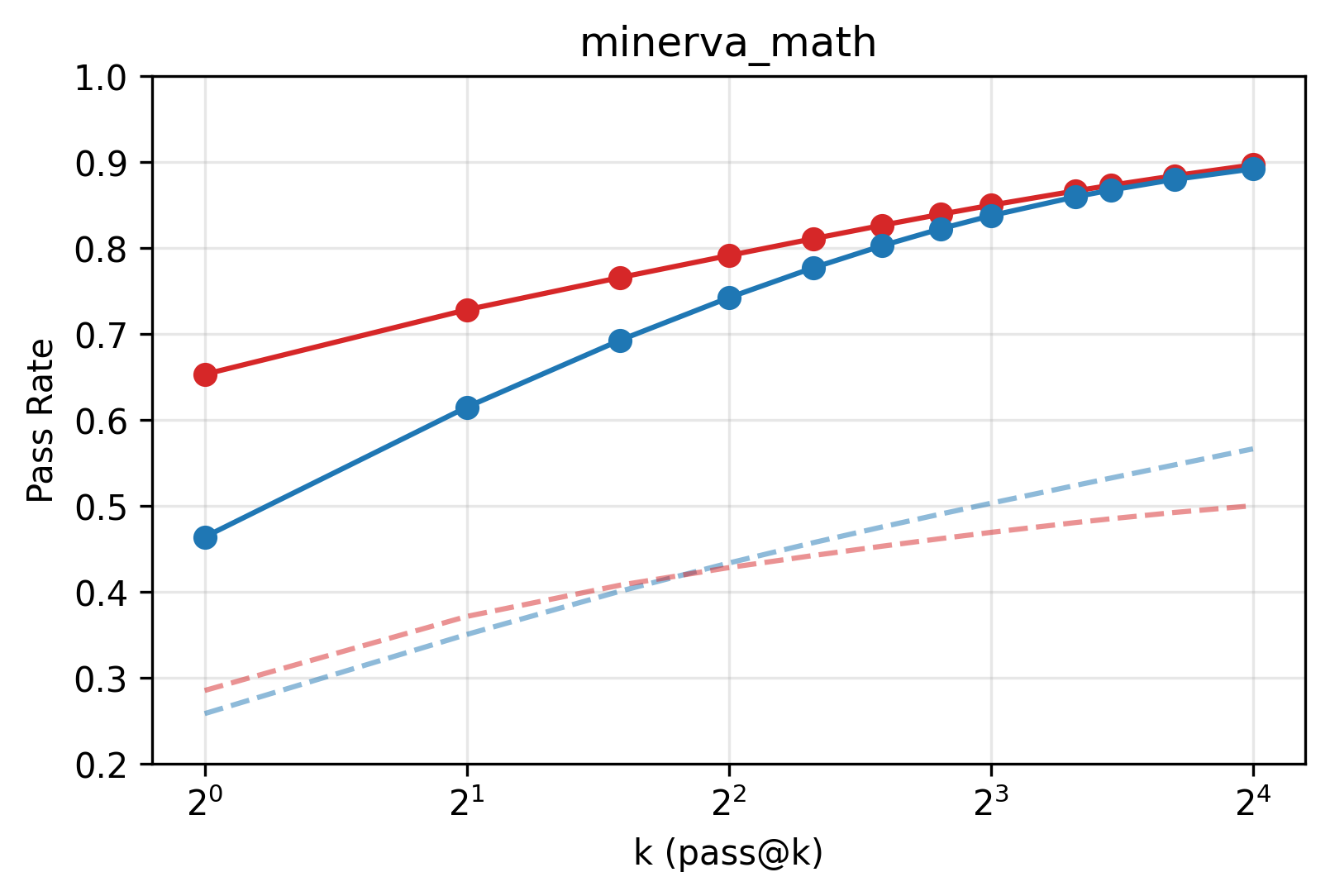} \\
\end{tabular}
\caption{Pass@k evaluation results comparing baseline evaluation methods (dashed line) with our LLM-as-a-judge evaluation approach (solid line) on the Qwen2.5-7B model. The left column shows results by the SimpleRL evaluation framework, while the right presents the Lighteval evaluation framework. Two datasets are presented: GSM8K (top) and Minerva (bottom). Our approach demonstrates consistent evaluation across both frameworks and outperforms baseline evaluation. Additional results for Qwen2.5-14B and 32B models are shown in the appendix (Figure~\ref{fig:passatk_7b_results_appendix}).}
\label{fig:passatk_7b_results1}
\end{figure*}

\subsection{Dataset Answer Validation}
\label{sec:methods:s2}

To validate the answer compared to the dataset GT, we provide the LLM judge with the question, along with the generated answer from the previous stage and the dataset GT answer (each question prompted independently). The LLM evaluates both answers and assesses the correctness of each. Finally, the LLM generates its own validated answer by synthesizing the observed answer candidates into a final concluded response. This answer will be used for the eventual LLM-as-a-judge evaluation of the responses. This approach enables us to obtain a finalized answer after considering both the dataset and the LLM answers, with the option to accept or reject each solution.

When the dataset GT answer is classified as incorrect or inconsistent, this may result from either erroneous dataset annotations or limitations in the validation process. As the underlying cause cannot always be reliably determined, we mark such cases as lacking evaluation applicability and exclude them from automatic evaluation. These cases constitute a small fraction of the dataset, and we favor improved evaluation reliability over full coverage. This filtering may slightly increase pass@k, as expected from removing ambiguous or invalid samples, and does not indicate improved model performance.

\subsection{Response Evaluation}
\label{sec:methods:s3}
The response evaluation is the main stage of our approach. We use LLM-as-a-judge to evaluate the model predictions' (mathematical answers) correctness by prompting with a mathematical question, the validated answer from the previous stage, and the model's predicted answers. The final answer of the mathematical reasoning is requested to be formatted inside a \textbackslash boxed\{\}, and we parse the final answer out of the full response. The judge LLM is requested by the prompt to classify each response as correct or incorrect. Due to the generative nature of language models, to achieve robust evaluation, we generate several predictions from a model and evaluate them all to compute the pass@k metric \cite{Chen2021EvaluatingLL}, which represents the distribution and variety of the model outputs.

The pass@k metric is formally defined as:
\begin{equation}
\text{pass@}k = \mathbb{E} \left[ 1 - \frac{\binom{n-c}{k}}{\binom{n}{k}} \right],
\end{equation}

where $n$ represents the total number of samples generated per question, $c$ denotes the number of correct samples for that question, and $k$ is the number of samples considered for evaluation. The expectation is computed over all questions in the evaluation set.
This formulation captures the likelihood of finding at least one correct answer when sampling $k$ responses from $n$ total generated responses, providing insight into both the model's accuracy and the effectiveness of generating multiple candidate solutions.

We evaluate multiple predictions simultaneously in each LLM call, rather than single predictions per prompt, to improve robustness by providing the model with broader context through diverse responses, following ablations and meta-evaluation for different group sizes (denoted by $n_{g}$).
Considering the set of unique answers from the parsed answers, $n=8$ samples are sampled and shuffled for the evaluation call by the LLM.
Random sampling and shuffling of the responses was applied to LLM prompting, which is important due to the positional bias in LLMs and LLM-as-a-judge specifically, where evaluations tend to be biased toward the input response position \cite{ye2024justice}. To improve robustness, each final answer was verified $n_{verif}$ times, and a majority decision was applied for the final correctness decision.

\subsection{Meta-Evaluation}
\label{sec:methods:meta-evaluation}

To evaluate our proposed approach, we conduct a \textit{meta-evaluation} \cite{li2024llms} to assess the accuracy and correctness of our evaluation pipeline.
Since we propose an improved evaluation method to address the insufficiencies of existing approaches, establishing a gold-standard benchmark evaluation becomes challenging and requires additional effort for numerical performance meta-evaluation.
The fundamental challenge lies in the open-ended nature of mathematically equivalent answer representations, since there is no closed set of all possible correct representations for a given answer, and comprehensive labeling is infeasible. To address this limitation, we create a small meta-evaluation dataset of responses from the Qwen2.5-7B model, which provides a finite set of predictions that can be systematically labeled. We randomly selected several questions from the dataset and manually annotated the correctness of model responses (640 responses). This process yields a closed set of labeled responses suitable for numerical meta-evaluation. We then compare the baseline symbolic evaluation, our proposed LLM-as-a-judge approach, and various design alternatives (ablations) using this annotated dataset.

\section{Results}
\label{sec:results}

\begin{table*}[t!]
\centering
\newcommand{\diffcolor}[1]{\textcolor{darkgreen}{\tiny #1}}
\newcommand{\setrowcolor}{\rowcolor[rgb]{ 0.973, 0.983, 0.99}}
\caption{\textbf{Pass@1 Evaluation Results.} Comparison of the baseline (symbolic) evaluation method with our LLM-as-a-judge approach across four mathematical reasoning datasets and three Qwen2.5 model sizes.}
\label{tab:passat1_results}
\small
\setlength{\tabcolsep}{4pt}
\begin{tabular}{lcccccccc}
\hline
& \multicolumn{2}{c}{GSM8K} & \multicolumn{2}{c}{Minerva} & \multicolumn{2}{c}{Math500} & \multicolumn{2}{c}{Olympiad} \\
Model & Baseline & Ours & Baseline & Ours & Baseline & Ours & Baseline & Ours \\
\hline
Qwen2.5-7B      & 85.0 & 86.8 \diffcolor{$\uparrow$1.8\%} & 23.0 & 47.4 \diffcolor{$\uparrow$24.4\%} & 61.5 & 63.2 \diffcolor{$\uparrow$1.7\%} & 28.1 & 30.6 \diffcolor{$\uparrow$2.5\%} \\
\setrowcolor ~~$\hookrightarrow$~+ SimpleRL & 92.2 & 93.8 \diffcolor{$\uparrow$1.6\%} & 35.1 & 65.7 \diffcolor{$\uparrow$30.6\%} & 78.0 & 79.3 \diffcolor{$\uparrow$1.3\%} & 41.2 & 42.7 \diffcolor{$\uparrow$1.5\%} \\
Qwen2.5-7B \tiny Lighteval      & 84.2 & 86.9 \diffcolor{$\uparrow$2.7\%} & 25.8 & 46.4 \diffcolor{$\uparrow$20.6\%} & 62.5 & 63.8 \diffcolor{$\uparrow$1.3\%} & 27.8 & 33.9 \diffcolor{$\uparrow$6.1\%} \\
\setrowcolor ~~$\hookrightarrow$~+ SimpleRL \tiny Lighteval & 71.9 & 93.0 \diffcolor{$\uparrow$21.1\%} & 28.5 & 65.3 \diffcolor{$\uparrow$36.8\%} & 56.2 & 77.9 \diffcolor{$\uparrow$21.7\%} & 27.1 & 46.1 \diffcolor{$\uparrow$19.0\%} \\

Qwen2.5-14B & 89.9 & 91.2 \diffcolor{$\uparrow$1.3\%} & 23.5 & 47.6 \diffcolor{$\uparrow$24.1\%} & 63.3 & 64.6 \diffcolor{$\uparrow$1.3\%} & 28.6 & 31.4 \diffcolor{$\uparrow$2.8\%} \\
\setrowcolor ~~$\hookrightarrow$~+ SimpleRL & 94.6 & 96.3 \diffcolor{$\uparrow$1.7\%} & 41.7 & 74.3 \diffcolor{$\uparrow$32.6\%} & 80.2 & 81.2 \diffcolor{$\uparrow$1.0\%} & 43.5 & 48.1 \diffcolor{$\uparrow$4.6\%} \\

Qwen2.5-32B & 90.7 & 93.0 \diffcolor{$\uparrow$2.3\%} & 27.5 & 56.0 \diffcolor{$\uparrow$28.5\%} & 64.7 & 66.4 \diffcolor{$\uparrow$1.7\%} & 27.9 & 30.6 \diffcolor{$\uparrow$2.7\%} \\
\setrowcolor ~~$\hookrightarrow$~+ SimpleRL & 95.9 & 97.8 \diffcolor{$\uparrow$1.9\%} & 43.5 & 74.8 \diffcolor{$\uparrow$31.3\%} & 82.5 & 84.2 \diffcolor{$\uparrow$1.7\%} & 46.0 & 52.0 \diffcolor{$\uparrow$6.0\%} \\
Llama3.1-8B & 44.3 & 45.4 \diffcolor{$\uparrow$1.1\%} & 5.4 & 9.4 \diffcolor{$\uparrow$4.0\%} & 12.9 & 13.6 \diffcolor{$\uparrow$0.7\%} & 2.7 & 3.5 \diffcolor{$\uparrow$0.8\%} \\

\hline
\end{tabular}
\end{table*}

\begin{table}[t!]
\centering
\caption{Example questions presenting the difference between symbolic and LLM-based evaluation approaches. Our approach generalizes to various answer representations and correctly evaluates responses where the symbolic evaluation fails.}
\label{tab:qexample1}
\setlength{\tabcolsep}{4pt}
\begin{tabular}{p{3cm} p{1.5cm} p{1.9cm}}
\hline
\multicolumn{3}{p{7.2cm}}{\small \textbf{Question:} "Suppose air molecules have a collision cross section of $10^{-16}$ cm$^2$. If the (number) density of air molecules is $10^{19}$ cm$^{-3}$, what is the collision mean free path in cm? Answer to one significant figure." \; \textcolor{navyblue}{GT: 1e-3} \; {\footnotesize (Minerva)} } \\
\hline
Answer & Symbolic & LLM \small (Ours) \\
\hline
\formatcorrecttext & & \\
\small\verb|0.001| & \Large \textcolor{darkgreen}{\cmark} & \Large \textcolor{darkgreen}{\cmark} \\
\small\verb|$10^{-3}$| & \Large \textcolor{red}{\xmark} & \Large \textcolor{darkgreen}{\cmark} \\
\formatincorrecttext & & \\
\small\verb|1| & \Large \textcolor{red}{\xmark} & \Large \textcolor{red}{\xmark} \\
\small\verb|0| & \Large \textcolor{red}{\xmark} & \Large \textcolor{red}{\xmark} \\
\small\verb|$10^{-35}$| & \Large \textcolor{red}{\xmark} & \Large \textcolor{red}{\xmark} \\
\small\verb|1000| & \Large \textcolor{red}{\xmark} & \Large \textcolor{red}{\xmark} \\
\small\verb|$2 \times 10^{12}$| & \Large \textcolor{red}{\xmark} & \Large \textcolor{red}{\xmark} \\
\small\verb|-35| & \Large \textcolor{red}{\xmark} & \Large \textcolor{red}{\xmark} \\
\hline\hline
\multicolumn{3}{p{7.2cm}}{\small \textbf{Question:} "Jordan wanted to surprise her mom with a homemade birthday cake. From reading the instructions, she knew it would take 20 minutes... , what is the latest time of day that Jordan can start making the cake to be ready to serve it at 5:00 pm?" \; \textcolor{navyblue}{GT: 2} \; {\footnotesize (GSM8K)} }  \\ %
\hline
Answer \small (Lighteval) & Symbolic & LLM \small (Ours) \\
\hline
\formatcorrecttext & & \\
\small\verb|2:00 pm| & \Large \textcolor{red}{\xmark} & \Large \textcolor{darkgreen}{\cmark} \\
\small\verb|2:00 \text{ pm}| & \Large \textcolor{red}{\xmark} & \Large \textcolor{darkgreen}{\cmark} \\
\small\verb|2:00 , \text{pm}| & \Large \textcolor{red}{\xmark} & \Large \textcolor{darkgreen}{\cmark} \\
\small\verb|2:00\ pm| & \Large \textcolor{red}{\xmark} & \Large \textcolor{darkgreen}{\cmark} \\
\formatincorrecttext & & \\
\small\verb|12:00 \text{ pm}| & \Large \textcolor{red}{\xmark} & \Large \textcolor{red}{\xmark} \\
\small\verb|12:40| & \Large \textcolor{red}{\xmark} & \Large \textcolor{red}{\xmark} \\
\hline
\end{tabular}
\vspace{-0.1in}
\end{table}

We compare our proposed LLM-as-a-judge evaluation against two commonly used evaluation pipelines: Lighteval \cite{lighteval}, a popular repository for LLM evaluations and mathematical benchmarks, and SimpleRL~\cite{zeng2025simplerl}, a framework focused on reinforcement learning for mathematical reasoning that includes mathematical evaluation on various benchmarks. Both baseline evaluations rely on symbolic mathematics packages (SymPy) to compare model predictions against GT answers.

We evaluate the Qwen2.5 model series \cite{qwen2024qwen2} across three sizes (7B, 14B, and 32B parameters) and the Llama3.1-8B model \cite{grattafiori2024llama}. Additionally, we test the RLVR versions of these models, which were trained using the SimpleRL framework with the GRPO algorithm \cite{shao2024deepseekmath} for enhanced mathematical reasoning capabilities. The pass@k metric evaluation results for the 7B model are presented in \Cref{fig:passatk_7b_results1}, comparing both SimpleRL and Lighteval frameworks, and pass@1 performance results are presented in \Cref{tab:passat1_results}. We observe consistent results of our evaluation for both frameworks, in contrast to the baselines that demonstrate different results between evaluations, which implies sensitivity to verification logic. Especially, for the RLVR trained model (Qwen2.5-7B-SimpleRL), Lighteval evaluation presents significant underestimation compared to the SimpleRL evaluation (which may contain a close match between training and evaluation performance). The performance consistency of our evaluation signifies the robustness of our approach across prediction and evaluation frameworks. Example questions and predictions are presented in \Cref{tab:examples1}, comparing the symbolic evaluation and our LLM-based evaluation. Additional examples and results for Qwen2.5 and Llama3.1-8B models and more datasets are shown in the Appendix (\Cref{fig:passatk_results_14b_32b,fig:passatk_7b_results_appendix,fig:lighteval_llama_results} and more).

\begin{figure}[tb!]
\centering
\newcommand{\si}{0.9}
\includegraphics[width=\si\linewidth]{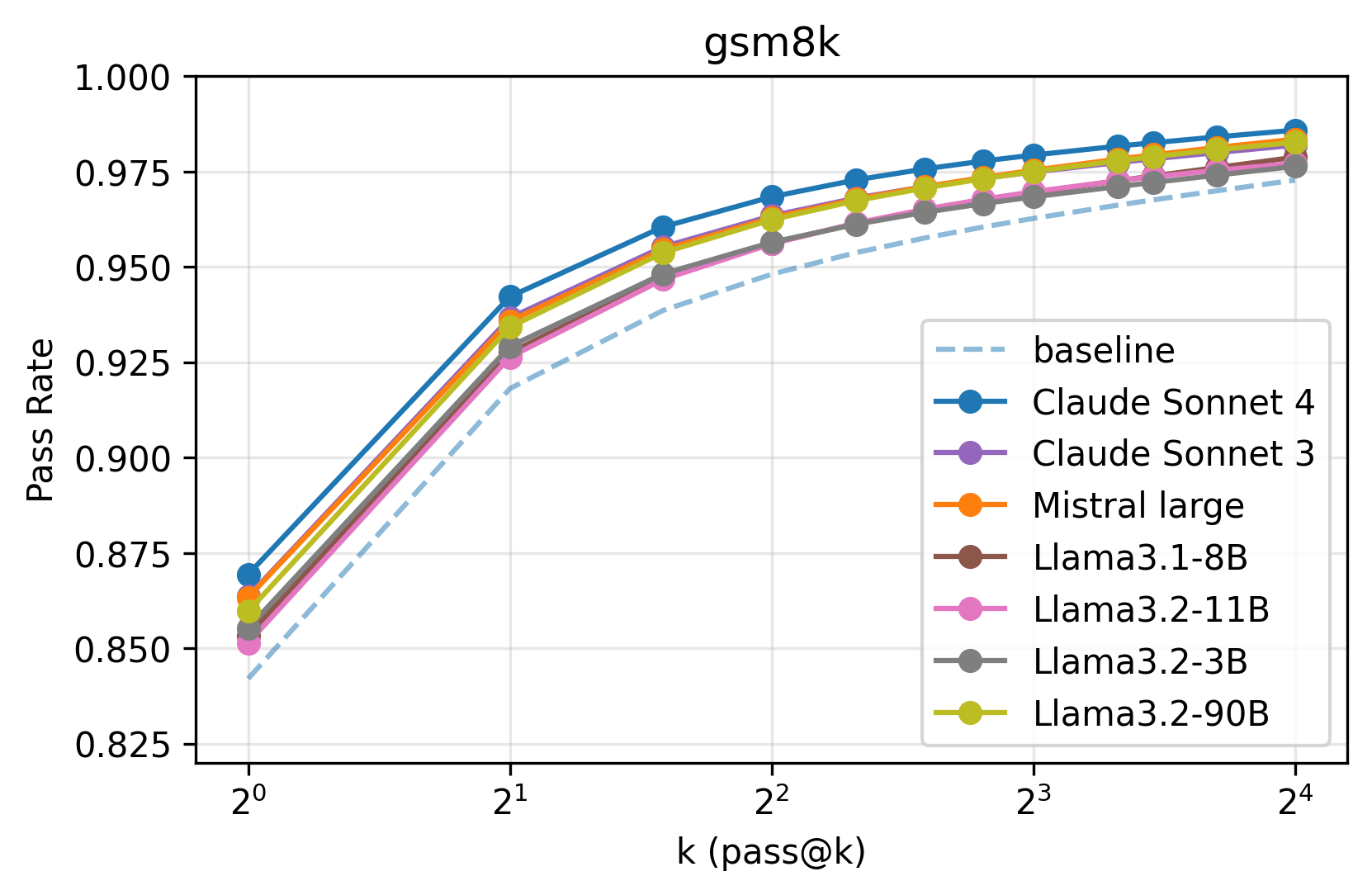}

\includegraphics[width=\si\linewidth]{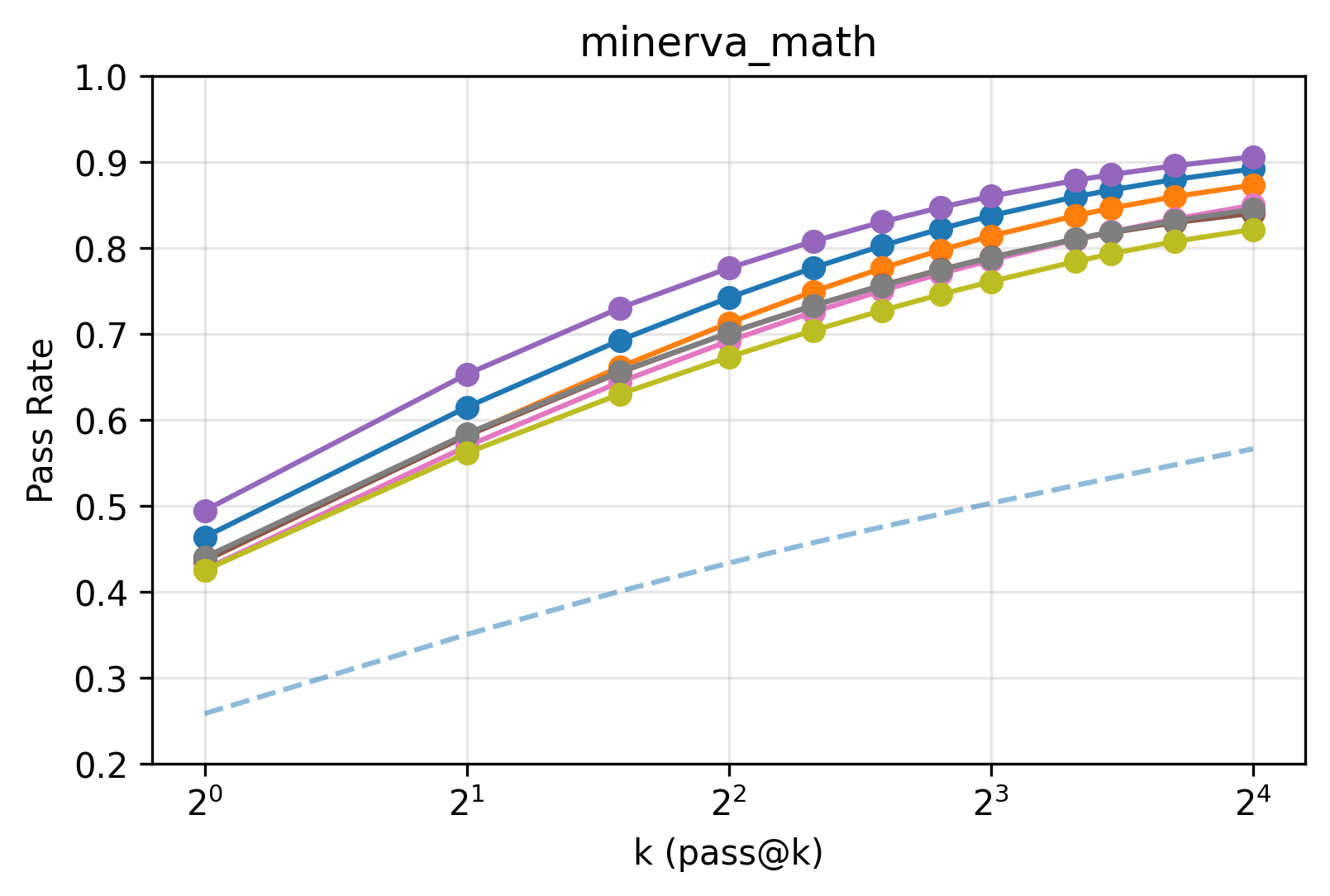}

\caption{Various judge models evaluation comparison (Claude, Llama, and Mistral). All judges present similar behavior and improved results over the baseline evaluation (SimpleRL, dashed line). Tested on Qwen2.5-7B predictions for GSM8K and Minerva datasets.}
\label{fig:multi_llm_comparison}
\vspace{-0.162in}
\end{figure}
\vspace{-0.05in}

\begin{table*}[h!]
\centering
\caption{\textbf{Meta-Evaluation and Ablation Results.} Numerical evaluation of the baseline (SimpleRL) performance and different configurations of our pipeline on a manually classified meta-evaluation dataset. Our pipeline significantly outperforms baseline evaluation {\small ($F_1=0.741$)}, while our proposed design achieves the best performance.}
\label{tab:ablations}
\small
\setlength{\tabcolsep}{4pt}
\begin{tabular}{lcccccc}
\hline
Configuration & Ans. Validation & Res. Evaluation & Group & Precision & Recall & F1 Score \\ [-2pt]
  & \footnotesize temperature & \footnotesize temperature & \footnotesize size $n_{g}$ &  &  &  \\
\hline
SimpleRL evaluation & -- & -- & -- & 0.989 & 0.592 & 0.741 \\
\hline
Ablations: \\
(a) & 0.1 & 0 & 8 & 0.925 & 0.783 & 0.848 \\
(b) & 0.1 & 0.1 & 8 & 0.935 & 0.783 & 0.852 \\
(c) & 0 & 0 & 8 & 0.940 & 0.919 & 0.929 \\
(d) & 0 & 0.1 & 1 & 0.957 & 0.905 & 0.930 \\
(e) & 0 & 0 & 1 & 0.957 & 0.905 & 0.930 \\
(f) & 0 & 0.1 & 4 & 0.906 & 1.000 & 0.951 \\
(g) & 0 & 0 & 4 & 0.932 & 0.986 & 0.958 \\
(h) \, $n_{verif}=5$ & 0 & 0.1 & 8 & 0.929 & 1.000 & 0.963 \\
(i) & -- & 0.1 & 8 & 0.929 & 1.000 & 0.963 \\
Proposed & 0 & 0.1 & 8 & 0.952 & 0.986 & \textbf{0.969} \\

\hline
\end{tabular}
\label{tab:bootstrapping_results}
\end{table*}

\vspace{-0.1in}

\subsection{LLM Judge Model}
We employed Claude-Sonnet-4 for our evaluations as the primary LLM judge model. In addition, we tested our evaluation pipeline using several LLMs: Claude-Sonnet-3, Mistral-large, and the Llama3 family (including smaller models), as presented in \Cref{fig:multi_llm_comparison}. All tested LLM judges demonstrated consistent behavior with minor variations between the models, specifically for the Minerva dataset, where the symbolic evaluation is underevaluating (as observed also in \Cref{tab:passat1_results}), which strengthens the stability of our approach and suggests the feasibility of using even smaller, more cost-effective LLMs for such an evaluation task for RLVR and benchmarking.

\subsection{Meta-Evaluation and Ablation Study}
The results, presented in \Cref{tab:ablations}, demonstrate the poor performance of baseline symbolic evaluation, with an F1 score of 0.741, compared to our LLM-as-a-judge approach, which demonstrates substantial improvement. A comprehensive ablation study validates our proposed configuration, achieving the highest F1 performance of 0.969.
We tested various group sizes ($n_{g}$) of responses for LLM queries (per question), particularly examining $n_{g}=1$ (configurations (d) and (e)) that show the contribution of group evaluation compared to individual evaluation ($n_{g}=8$). When using an LLM temperature of 0.1 during the answer validation stage, we observed significant performance degradation (configurations (a) and (b)); therefore, we adopted a temperature of 0 for this stage.
We also evaluated each response 5 times ($n_{verif}=5$, (configuration (h)) with temperature of 0.1 with majority voting, but observed no significant improvement. Consequently, we maintain $n_{verif}=3$ in our proposed approach to avoid redundant computations. Testing a single-stage GT answer validation approach (config (i)) without an independent answering stage resulted in a modest performance decrease.

\section{Discussion}
\label{sec:discussion}
Our experimental results demonstrate that the proposed LLM-as-a-judge evaluation framework significantly mitigates the flaws of the traditional symbolic verification method and improves the evaluation quality across multiple mathematical reasoning benchmarks and models. The consistent improvements observed across frameworks and judge models validate the robustness of our approach, as confirmed by the meta-evaluation and ablation study we conducted. We demonstrated that various and even smaller LLMs obtain similar results, demonstrating the broad applicability of the method.

This work addresses a critical bottleneck in mathematical reasoning evaluation with implications for model intelligence assessment and current evaluation reliability. Our improved evaluation framework also has a great benefit on training methods that are based on verifiable rewards, such as RLVR, where evaluation correctness affects model optimization and overall performance.

\clearpage
\section*{Limitations}
\label{sec:limitations}
While our approach demonstrates significant improvements over symbolic evaluation methods, several limitations should be acknowledged. First, LLM-based evaluation requires substantially more computational resources compared to symbolic verification, as each question evaluation involves multiple LLM calls, making it more expensive for large-scale benchmarking scenarios. Additionally, the evaluation quality is inherently dependent on the capabilities and potential biases of the LLM judge. While we tested multiple LLMs and observed consistent behavior, small differences were still observed, indicating that the evaluation is subject to the limitations and potential hallucinations of the judge model.

We acknowledge that the meta-evaluation was conducted on a small dataset due to manual annotation and expert knowledge constraints. While this meta-evaluation provides valuable validation for our suggested approach, a larger-scale meta-evaluation can be performed to strengthen our claims.

Since the model is required to provide the final answer in a \textbackslash boxed\{\} format, we rely on the assumption of correct formatting. If the model does not respond in the correct format, identifying the final answer might be challenging, though an LLM can be used for this parsing task as well to generalize even further.

While our method handles diverse representations better than symbolic approaches, it still relies on the LLM's ability to parse and understand mathematical notation, which may fail for highly specialized or non-standard mathematical expressions. Despite using multiple verification and majority voting, there may still be inconsistencies in evaluation decisions, particularly for borderline cases or when the mathematical correctness is ambiguous.

\bibliography{references}

\begin{thebibliography}{37}
\providecommand{\natexlab}[1]{#1}

\bibitem[{Chen et~al.(2021)Chen, Tworek, Jun, Yuan, Pond{\'e}, Kaplan, Edwards, Burda, Joseph, Brockman, Ray, Puri, Krueger, Petrov, Khlaaf, Sastry, Mishkin, Chan, Gray, Ryder, Pavlov, Power, Kaiser, Bavarian, Winter, Tillet, Such, Cummings, Plappert, Chantzis, Barnes, Herbert-Voss, Guss, Nichol, Babuschkin, Balaji, Jain, Carr, Leike, Achiam, Misra, Morikawa, Radford, Knight, Brundage, Murati, Mayer, Welinder, McGrew, Amodei, McCandlish, Sutskever, and Zaremba}]{Chen2021EvaluatingLL}
Mark Chen, Jerry Tworek, Heewoo Jun, Qiming Yuan, Henrique Pond{\'e}, Jared Kaplan, Harrison Edwards, Yura Burda, Nicholas Joseph, Greg Brockman, Alex Ray, Raul Puri, Gretchen Krueger, Michael Petrov, Heidy Khlaaf, Girish Sastry, Pamela Mishkin, Brooke Chan, Scott Gray, and 34 others. 2021.
\newblock \href {https://api.semanticscholar.org/CorpusID:235755472} {Evaluating large language models trained on code}.
\newblock \emph{ArXiv}, abs/2107.03374.

\bibitem[{Cobbe et~al.(2021)Cobbe, Kosaraju, Bavarian, Chen, Jun, Kaiser, Plappert, Tworek, Hilton, Nakano et~al.}]{cobbe2021training}
Karl Cobbe, Vineet Kosaraju, Mohammad Bavarian, Mark Chen, Heewoo Jun, Lukasz Kaiser, Matthias Plappert, Jerry Tworek, Jacob Hilton, Reiichiro Nakano, and 1 others. 2021.
\newblock Training verifiers to solve math word problems.
\newblock \emph{arXiv preprint arXiv:2110.14168}.

\bibitem[{Gao et~al.(2024)Gao, Song, Yang, Cai, Miao, Dong, Li, Ma, Chen, Xu et~al.}]{gao2024omni}
Bofei Gao, Feifan Song, Zhe Yang, Zefan Cai, Yibo Miao, Qingxiu Dong, Lei Li, Chenghao Ma, Liang Chen, Runxin Xu, and 1 others. 2024.
\newblock Omni-math: A universal olympiad level mathematic benchmark for large language models.
\newblock \emph{arXiv preprint arXiv:2410.07985}.

\bibitem[{Gema et~al.(2025)Gema, Leang, Hong, Devoto, Mancino, Saxena, He, Zhao, Du, Madani et~al.}]{gema2025we}
Aryo~Pradipta Gema, Joshua Ong~Jun Leang, Giwon Hong, Alessio Devoto, Alberto Carlo~Maria Mancino, Rohit Saxena, Xuanli He, Yu~Zhao, Xiaotang Du, Mohammad Reza~Ghasemi Madani, and 1 others. 2025.
\newblock Are we done with mmlu?
\newblock In \emph{Proceedings of the 2025 Conference of the Nations of the Americas Chapter of the Association for Computational Linguistics: Human Language Technologies (Volume 1: Long Papers)}, pages 5069--5096.

\bibitem[{Grattafiori et~al.(2024)Grattafiori, Dubey, Jauhri, Pandey, Kadian, Al-Dahle, Letman, Mathur, Schelten, Vaughan et~al.}]{grattafiori2024llama}
Aaron Grattafiori, Abhimanyu Dubey, Abhinav Jauhri, Abhinav Pandey, Abhishek Kadian, Ahmad Al-Dahle, Aiesha Letman, Akhil Mathur, Alan Schelten, Alex Vaughan, and 1 others. 2024.
\newblock The llama 3 herd of models.
\newblock \emph{arXiv preprint arXiv:2407.21783}.

\bibitem[{Gu et~al.(2024)Gu, Jiang, Shi, Tan, Zhai, Xu, Li, Shen, Ma, Liu et~al.}]{gu2024survey}
Jiawei Gu, Xuhui Jiang, Zhichao Shi, Hexiang Tan, Xuehao Zhai, Chengjin Xu, Wei Li, Yinghan Shen, Shengjie Ma, Honghao Liu, and 1 others. 2024.
\newblock A survey on llm-as-a-judge.
\newblock \emph{arXiv preprint arXiv:2411.15594}.

\bibitem[{Guo et~al.(2025)Guo, Yang, Zhang, Song, Zhang, Xu, Zhu, Ma, Wang, Bi et~al.}]{guo2025deepseek}
Daya Guo, Dejian Yang, Haowei Zhang, Junxiao Song, Ruoyu Zhang, Runxin Xu, Qihao Zhu, Shirong Ma, Peiyi Wang, Xiao Bi, and 1 others. 2025.
\newblock Deepseek-r1: Incentivizing reasoning capability in llms via reinforcement learning.
\newblock \emph{arXiv preprint arXiv:2501.12948}.

\bibitem[{Habib et~al.(2023)Habib, Fourrier, Kydlíček, Wolf, and Tunstall}]{lighteval}
Nathan Habib, Clémentine Fourrier, Hynek Kydlíček, Thomas Wolf, and Lewis Tunstall. 2023.
\newblock \href {https://github.com/huggingface/lighteval} {Lighteval: A lightweight framework for llm evaluation}.

\bibitem[{He et~al.(2024)He, Luo, Bai, Hu, Thai, Shen, Hu, Han, Huang, Zhang et~al.}]{he2024olympiadbench}
Chaoqun He, Renjie Luo, Yuzhuo Bai, Shengding Hu, Zhen Thai, Junhao Shen, Jinyi Hu, Xu~Han, Yujie Huang, Yuxiang Zhang, and 1 others. 2024.
\newblock Olympiadbench: A challenging benchmark for promoting agi with olympiad-level bilingual multimodal scientific problems.
\newblock In \emph{Proceedings of the 62nd Annual Meeting of the Association for Computational Linguistics (Volume 1: Long Papers)}, pages 3828--3850.

\bibitem[{Hendrycks et~al.(2021)Hendrycks, Burns, Kadavath, Arora, Basart, Tang, Song, and Steinhardt}]{hendrycks2021measuring}
Dan Hendrycks, Collin Burns, Saurav Kadavath, Akul Arora, Steven Basart, Eric Tang, Dawn Song, and Jacob Steinhardt. 2021.
\newblock Measuring mathematical problem solving with the math dataset.
\newblock \emph{arXiv preprint arXiv:2103.03874}.

\bibitem[{Lambert et~al.(2024)Lambert, Morrison, Pyatkin, Huang, Ivison, Brahman, Miranda, Liu, Dziri, Lyu et~al.}]{lambert2024tulu}
Nathan Lambert, Jacob Morrison, Valentina Pyatkin, Shengyi Huang, Hamish Ivison, Faeze Brahman, Lester James~V Miranda, Alisa Liu, Nouha Dziri, Shane Lyu, and 1 others. 2024.
\newblock Tulu 3: Pushing frontiers in open language model post-training.
\newblock \emph{arXiv preprint arXiv:2411.15124}.

\bibitem[{Lewkowycz et~al.(2022)Lewkowycz, Andreassen, Dohan, Dyer, Michalewski, Ramasesh, Slone, Anil, Schlag, Gutman-Solo et~al.}]{lewkowycz2022solving}
Aitor Lewkowycz, Anders Andreassen, David Dohan, Ethan Dyer, Henryk Michalewski, Vinay Ramasesh, Ambrose Slone, Cem Anil, Imanol Schlag, Theo Gutman-Solo, and 1 others. 2022.
\newblock Solving quantitative reasoning problems with language models.
\newblock \emph{Advances in neural information processing systems}, 35:3843--3857.

\bibitem[{Li et~al.(2024{\natexlab{a}})Li, Dong, Chen, Su, Zhou, Ai, Ye, and Liu}]{li2024llms}
Haitao Li, Qian Dong, Junjie Chen, Huixue Su, Yujia Zhou, Qingyao Ai, Ziyi Ye, and Yiqun Liu. 2024{\natexlab{a}}.
\newblock Llms-as-judges: a comprehensive survey on llm-based evaluation methods.
\newblock \emph{arXiv preprint arXiv:2412.05579}.

\bibitem[{Li et~al.(2024{\natexlab{b}})Li, Chiang, Frick, Dunlap, Wu, Zhu, Gonzalez, and Stoica}]{li2024crowdsourced}
Tianle Li, Wei-Lin Chiang, Evan Frick, Lisa Dunlap, Tianhao Wu, Banghua Zhu, Joseph~E Gonzalez, and Ion Stoica. 2024{\natexlab{b}}.
\newblock From crowdsourced data to high-quality benchmarks: Arena-hard and benchbuilder pipeline.
\newblock \emph{arXiv preprint arXiv:2406.11939}.

\bibitem[{Li et~al.(2024{\natexlab{c}})Li, Wang, Li, Guo, Zhang, and Feng}]{li2024evaluating}
Xiaoyuan Li, Wenjie Wang, Moxin Li, Junrong Guo, Yang Zhang, and Fuli Feng. 2024{\natexlab{c}}.
\newblock Evaluating mathematical reasoning of large language models: A focus on error identification and correction.
\newblock \emph{arXiv preprint arXiv:2406.00755}.

\bibitem[{Lightman et~al.(2023)Lightman, Kosaraju, Burda, Edwards, Baker, Lee, Leike, Schulman, Sutskever, and Cobbe}]{lightman2023let}
Hunter Lightman, Vineet Kosaraju, Yuri Burda, Harrison Edwards, Bowen Baker, Teddy Lee, Jan Leike, John Schulman, Ilya Sutskever, and Karl Cobbe. 2023.
\newblock Let's verify step by step.
\newblock In \emph{The Twelfth International Conference on Learning Representations}.

\bibitem[{Liu et~al.(2025)Liu, Chen, Fang, Luo, Tian, and Liu}]{liu2025matheval}
Tianqiao Liu, Zui Chen, Zhensheng Fang, Weiqi Luo, Mi~Tian, and Zitao Liu. 2025.
\newblock Matheval: A comprehensive benchmark for evaluating large language models on mathematical reasoning capabilities.
\newblock \emph{Frontiers of Digital Education}, 2(2):16.

\bibitem[{Meurer et~al.(2017)Meurer, Smith, Paprocki, \v{C}ert\'{i}k, Kirpichev, Rocklin, Kumar, Ivanov, Moore, Singh, Rathnayake, Vig, Granger, Muller, Bonazzi, Gupta, Vats, Johansson, Pedregosa, Curry, Terrel, Rou\v{c}ka, Saboo, Fernando, Kulal, Cimrman, and Scopatz}]{sympy}
Aaron Meurer, Christopher~P. Smith, Mateusz Paprocki, Ond\v{r}ej \v{C}ert\'{i}k, Sergey~B. Kirpichev, Matthew Rocklin, AMiT Kumar, Sergiu Ivanov, Jason~K. Moore, Sartaj Singh, Thilina Rathnayake, Sean Vig, Brian~E. Granger, Richard~P. Muller, Francesco Bonazzi, Harsh Gupta, Shivam Vats, Fredrik Johansson, Fabian Pedregosa, and 8 others. 2017.
\newblock \href {https://doi.org/10.7717/peerj-cs.103} {Sympy: symbolic computing in python}.
\newblock \emph{PeerJ Computer Science}, 3:e103.

\bibitem[{Mirzadeh et~al.(2024)Mirzadeh, Alizadeh, Shahrokhi, Tuzel, Bengio, and Farajtabar}]{mirzadeh2024gsm}
Iman Mirzadeh, Keivan Alizadeh, Hooman Shahrokhi, Oncel Tuzel, Samy Bengio, and Mehrdad Farajtabar. 2024.
\newblock Gsm-symbolic: Understanding the limitations of mathematical reasoning in large language models.
\newblock \emph{arXiv preprint arXiv:2410.05229}.

\bibitem[{Qwen et~al.(2024)Qwen, Yang, Zhang, Hui, Zheng, Yu, Li, Liu, Huang, Wei et~al.}]{qwen2024qwen2}
A~Yang Qwen, Baosong Yang, B~Zhang, B~Hui, B~Zheng, B~Yu, Chengpeng Li, D~Liu, F~Huang, H~Wei, and 1 others. 2024.
\newblock Qwen2. 5 technical report.
\newblock \emph{arXiv preprint}.

\bibitem[{Shao et~al.(2024)Shao, Wang, Zhu, Xu, Song, Bi, Zhang, Zhang, Li, Wu et~al.}]{shao2024deepseekmath}
Zhihong Shao, Peiyi Wang, Qihao Zhu, Runxin Xu, Junxiao Song, Xiao Bi, Haowei Zhang, Mingchuan Zhang, YK~Li, Yang Wu, and 1 others. 2024.
\newblock Deepseekmath: Pushing the limits of mathematical reasoning in open language models.
\newblock \emph{arXiv preprint arXiv:2402.03300}.

\bibitem[{Shen et~al.(2025)Shen, Wong, He, Liang, Qiang, Meng, Zhao, Zeng, Zhu, Cui et~al.}]{shen2025let}
Chengyu Shen, Zhen~Hao Wong, Runming He, Hao Liang, Meiyi Qiang, Zimo Meng, Zhengyang Zhao, Bohan Zeng, Zhengzhou Zhu, Bin Cui, and 1 others. 2025.
\newblock Let's verify math questions step by step.
\newblock \emph{arXiv preprint arXiv:2505.13903}.

\bibitem[{Stephan et~al.(2024)Stephan, Zhu, A{\ss}enmacher, Shen, and Roth}]{stephan2024calculation}
Andreas Stephan, Dawei Zhu, Matthias A{\ss}enmacher, Xiaoyu Shen, and Benjamin Roth. 2024.
\newblock From calculation to adjudication: Examining llm judges on mathematical reasoning tasks.
\newblock \emph{arXiv preprint arXiv:2409.04168}.

\bibitem[{Tan et~al.(2024{\natexlab{a}})Tan, Zhuang, Montgomery, Tang, Cuadron, Wang, Popa, and Stoica}]{tan2024judgebench}
Sijun Tan, Siyuan Zhuang, Kyle Montgomery, William~Y Tang, Alejandro Cuadron, Chenguang Wang, Raluca~Ada Popa, and Ion Stoica. 2024{\natexlab{a}}.
\newblock Judgebench: A benchmark for evaluating llm-based judges.
\newblock \emph{arXiv preprint arXiv:2410.12784}.

\bibitem[{Tan et~al.(2024{\natexlab{b}})Tan, Li, Wang, Beigi, Jiang, Bhattacharjee, Karami, Li, Cheng, and Liu}]{tan2024large}
Zhen Tan, Dawei Li, Song Wang, Alimohammad Beigi, Bohan Jiang, Amrita Bhattacharjee, Mansooreh Karami, Jundong Li, Lu~Cheng, and Huan Liu. 2024{\natexlab{b}}.
\newblock Large language models for data annotation and synthesis: A survey.
\newblock \emph{arXiv preprint arXiv:2402.13446}.

\bibitem[{Vendrow et~al.(2025)Vendrow, Vendrow, Beery, and Madry}]{vendrow2025large}
Joshua Vendrow, Edward Vendrow, Sara Beery, and Aleksander Madry. 2025.
\newblock Do large language model benchmarks test reliability?
\newblock \emph{arXiv preprint arXiv:2502.03461}.

\bibitem[{Wang et~al.(2024)Wang, Li, Shao, Xu, Dai, Li, Chen, Wu, and Sui}]{wang2024math}
Peiyi Wang, Lei Li, Zhihong Shao, Runxin Xu, Damai Dai, Yifei Li, Deli Chen, Yu~Wu, and Zhifang Sui. 2024.
\newblock Math-shepherd: Verify and reinforce llms step-by-step without human annotations.
\newblock In \emph{Proceedings of the 62nd Annual Meeting of the Association for Computational Linguistics (Volume 1: Long Papers)}, pages 9426--9439.

\bibitem[{Wang et~al.(2022)Wang, Wei, Schuurmans, Le, Chi, Narang, Chowdhery, and Zhou}]{wang2022self}
Xuezhi Wang, Jason Wei, Dale Schuurmans, Quoc Le, Ed~Chi, Sharan Narang, Aakanksha Chowdhery, and Denny Zhou. 2022.
\newblock Self-consistency improves chain of thought reasoning in language models.
\newblock \emph{arXiv preprint arXiv:2203.11171}.

\bibitem[{Wei et~al.(2022)Wei, Wang, Schuurmans, Bosma, Xia, Chi, Le, Zhou et~al.}]{wei2022chain}
Jason Wei, Xuezhi Wang, Dale Schuurmans, Maarten Bosma, Fei Xia, Ed~Chi, Quoc~V Le, Denny Zhou, and 1 others. 2022.
\newblock Chain-of-thought prompting elicits reasoning in large language models.
\newblock \emph{Advances in neural information processing systems}, 35:24824--24837.

\bibitem[{Wu et~al.(2024)Wu, Yuan, Golovneva, Xu, Tian, Jiao, Weston, and Sukhbaatar}]{wu2024meta}
Tianhao Wu, Weizhe Yuan, Olga Golovneva, Jing Xu, Yuandong Tian, Jiantao Jiao, Jason Weston, and Sainbayar Sukhbaatar. 2024.
\newblock Meta-rewarding language models: Self-improving alignment with llm-as-a-meta-judge.
\newblock \emph{arXiv preprint arXiv:2407.19594}.

\bibitem[{Xia et~al.(2025)Xia, Li, Liu, Wu, and Liu}]{xia2025evaluating}
Shijie Xia, Xuefeng Li, Yixin Liu, Tongshuang Wu, and Pengfei Liu. 2025.
\newblock Evaluating mathematical reasoning beyond accuracy.
\newblock In \emph{Proceedings of the AAAI Conference on Artificial Intelligence}, volume~39, pages 27723--27730.

\bibitem[{Yang et~al.(2024)Yang, Zhang, Hui, Gao, Yu, Li, Liu, Tu, Zhou, Lin et~al.}]{yang2024qwen2}
An~Yang, Beichen Zhang, Binyuan Hui, Bofei Gao, Bowen Yu, Chengpeng Li, Dayiheng Liu, Jianhong Tu, Jingren Zhou, Junyang Lin, and 1 others. 2024.
\newblock Qwen2. 5-math technical report: Toward mathematical expert model via self-improvement.
\newblock \emph{arXiv preprint arXiv:2409.12122}.

\bibitem[{Ye et~al.(2024)Ye, Wang, Huang, Chen, Zhang, Moniz, Gao, Geyer, Huang, Chen et~al.}]{ye2024justice}
Jiayi Ye, Yanbo Wang, Yue Huang, Dongping Chen, Qihui Zhang, Nuno Moniz, Tian Gao, Werner Geyer, Chao Huang, Pin-Yu Chen, and 1 others. 2024.
\newblock Justice or prejudice? quantifying biases in llm-as-a-judge.
\newblock \emph{arXiv preprint arXiv:2410.02736}.

\bibitem[{Yue et~al.(2025)Yue, Chen, Lu, Zhao, Wang, Song, and Huang}]{yue2025does}
Yang Yue, Zhiqi Chen, Rui Lu, Andrew Zhao, Zhaokai Wang, Shiji Song, and Gao Huang. 2025.
\newblock Does reinforcement learning really incentivize reasoning capacity in llms beyond the base model?
\newblock \emph{arXiv preprint arXiv:2504.13837}.

\bibitem[{Zeng et~al.(2025)Zeng, Huang, Liu, Liu, He, Ma, and He}]{zeng2025simplerl}
Weihao Zeng, Yuzhen Huang, Qian Liu, Wei Liu, Keqing He, Zejun Ma, and Junxian He. 2025.
\newblock Simplerl-zoo: Investigating and taming zero reinforcement learning for open base models in the wild.
\newblock \emph{arXiv preprint arXiv:2503.18892}.

\bibitem[{Zheng et~al.(2025)Zheng, Zhang, Zhang, Lin, Lu, Yu, Liu, Zhou, and Lin}]{zheng2025processbench}
Chujie Zheng, Zhenru Zhang, Beichen Zhang, Runji Lin, Keming Lu, Bowen Yu, Dayiheng Liu, Jingren Zhou, and Junyang Lin. 2025.
\newblock Processbench: Identifying process errors in mathematical reasoning.
\newblock In \emph{Proceedings of the 63rd Annual Meeting of the Association for Computational Linguistics (Volume 1: Long Papers)}, pages 1009--1024.

\bibitem[{Zheng et~al.(2023)Zheng, Chiang, Sheng, Zhuang, Wu, Zhuang, Lin, Li, Li, Xing et~al.}]{zheng2023judging}
Lianmin Zheng, Wei-Lin Chiang, Ying Sheng, Siyuan Zhuang, Zhanghao Wu, Yonghao Zhuang, Zi~Lin, Zhuohan Li, Dacheng Li, Eric Xing, and 1 others. 2023.
\newblock Judging llm-as-a-judge with mt-bench and chatbot arena.
\newblock \emph{Advances in neural information processing systems}, 36:46595--46623.

\end{thebibliography}
\clearpage

\appendix

\definecolor{dark_erez}{RGB}{134,0,55}

\newcommand{\emphtxt}[1]{\textbf{{\color{dark_erez}#1}}}
\section*{Appendix}
\section{Terminology Conventions}
\label{appendix:terminology}

To ensure clarity throughout this paper, we adopt the following terminology conventions to distinguish between different assessment processes in our framework:

\paragraph{Evaluation.} Refers to the overall process of assessing model performance on mathematical reasoning benchmarks. This encompasses the pipeline of evaluating the predictions and measuring how well a language model performs on a given dataset.

\paragraph{Verification.} Refers specifically to the process of comparing a model-generated response against a GT answer to determine correctness. This is the core operation performed by our LLM-as-a-judge approach, where the judge determines whether a predicted answer is mathematically equivalent to the reference answer, in contrast to the symbolic evaluation.

\paragraph{Validation.} Refers to the preliminary step of confirming the correctness and consistency of the dataset's GT answers before using them for verification. This step addresses potential errors or ambiguities in benchmark annotations, ensuring that model predictions are compared against reliable reference answers.

\paragraph{Meta-evaluation.} Refers to the process of assessing the quality and accuracy of the evaluation framework itself. This involves comparing the judgments produced by our LLM-based verification against human-annotated correctness labels to quantify the reliability of our approach.

These terms are primarily used in the contexts defined above, though they may also appear elsewhere in the text with their standard meaning, separate from the specific processes described in this section.

\section{Question Clarity Analysis}
\label{appendix:calrity_score}
As described in \Cref{sec:methods:s1}, our framework generates a clarity score for each question to identify potentially problematic or ambiguous problems in the benchmark datasets. Questions with low clarity scores often contain logical inconsistencies, missing information, or ambiguous references that make definitive evaluation unreliable.

\Cref{tab:unclear_questions} presents representative examples of unclear questions identified in the GSM8K dataset, each receiving a clarity score of 3 or below with our LLM-based approach. These examples illustrate common issues, including logical contradictions (e.g., stating that all items are sold while asking how many remain), numerical inconsistencies (e.g., conflicting counts of stalls), and undefined references (e.g., doubling an unspecified quantity).

To ensure evaluation reliability, we filter out questions with clarity scores of 3 or below from our analysis. This dataset cleaning eliminates problematic questions where even human evaluators might disagree on the correct interpretation, thereby providing a more accurate assessment of both model performance and our evaluation framework's effectiveness.

\begin{table*}[t]
\centering
\definecolor{marktxtcolor}{RGB}{255,0,0}
\caption{Examples of unclear math questions from the GSM8K dataset identified by our LLM-based evaluation framework. These math questions have a clarity score of 3.}
\label{tab:unclear_questions}
\setlength{\tabcolsep}{4pt}
\begin{tabular}{p{14cm}}
\hline
Question: \; \small
"John plans to sell \textcolor{marktxtcolor}{all} his toys and use the money to buy video games. He has 13 lego sets and he sells them for \$15 each. He ends up buying 8 video games for \$20 each and has \$5 left. How many lego sets does he \textcolor{marktxtcolor}{still have}?" \\
\textit{Logical inconsistency: states John sells all toys but asks how many he still has.} \\
\hline
Question: \; \small
"\textcolor{marktxtcolor}{Ten stalls} have 20 cows each. Mr. Sylas buys 40 cows and divides them equally, putting an equal number of the new cows into each of the \textcolor{marktxtcolor}{twenty stalls}. How many cows are in 8 of the stalls?" \\
\textit{Inconsistency: mentions both "ten stalls" and "twenty stalls".} \\
\hline
Question: \; \small
"Wendy wants to place 20 more than \textcolor{marktxtcolor}{double the number of books} in a shelving system with 6 rows and 6 columns. How many books will she need to carry to complete her task?" \\
\textit{Unclear reference: "double the number of books" lacks baseline specification.} \\
\hline
\end{tabular}
\end{table*}

\begin{figure*}[t]
\centering
\newcommand{\plotwidthllama}{0.4\textwidth}
\begin{tabular}{cc}
\includegraphics[width=\plotwidthllama]{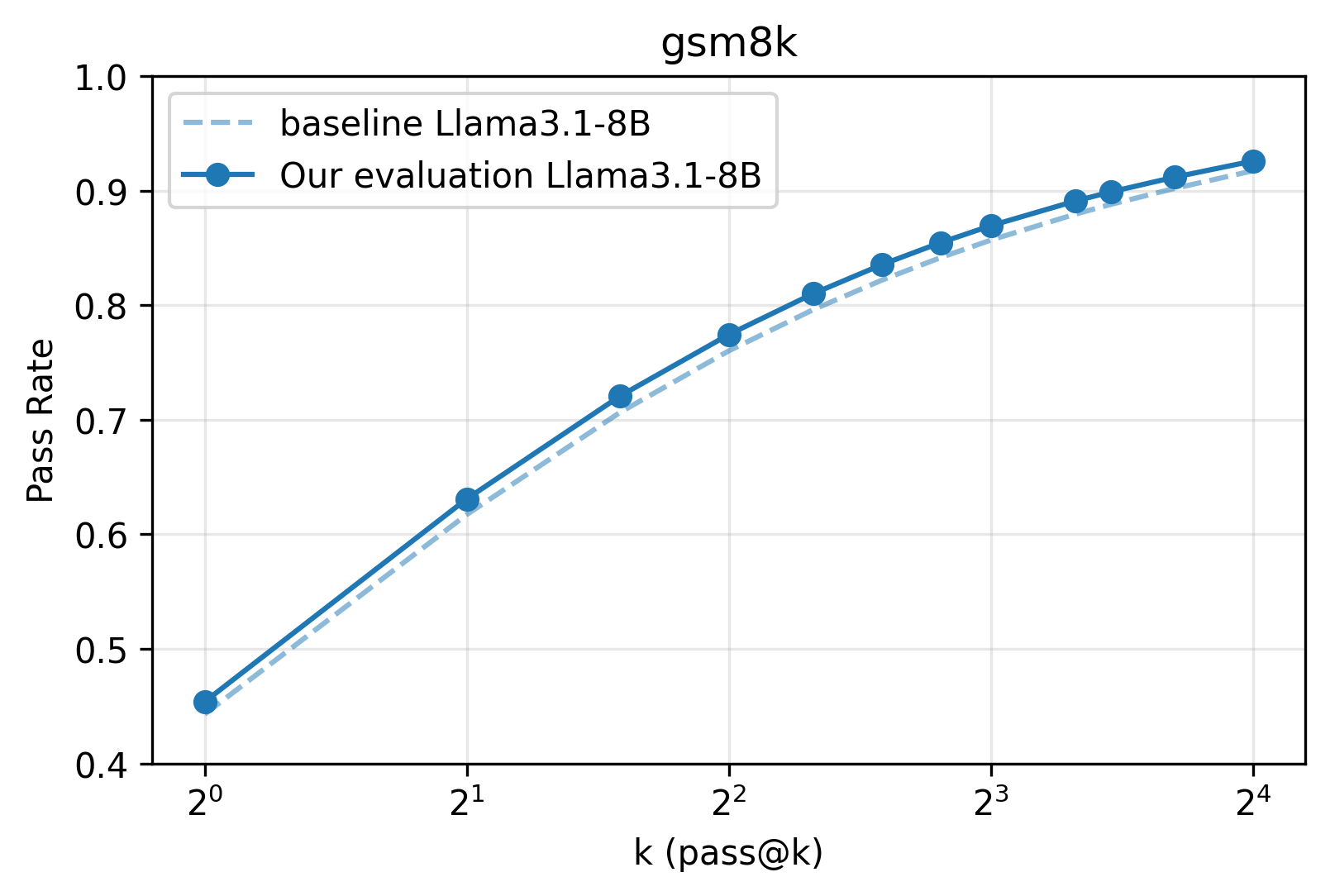} &
\includegraphics[width=\plotwidthllama]{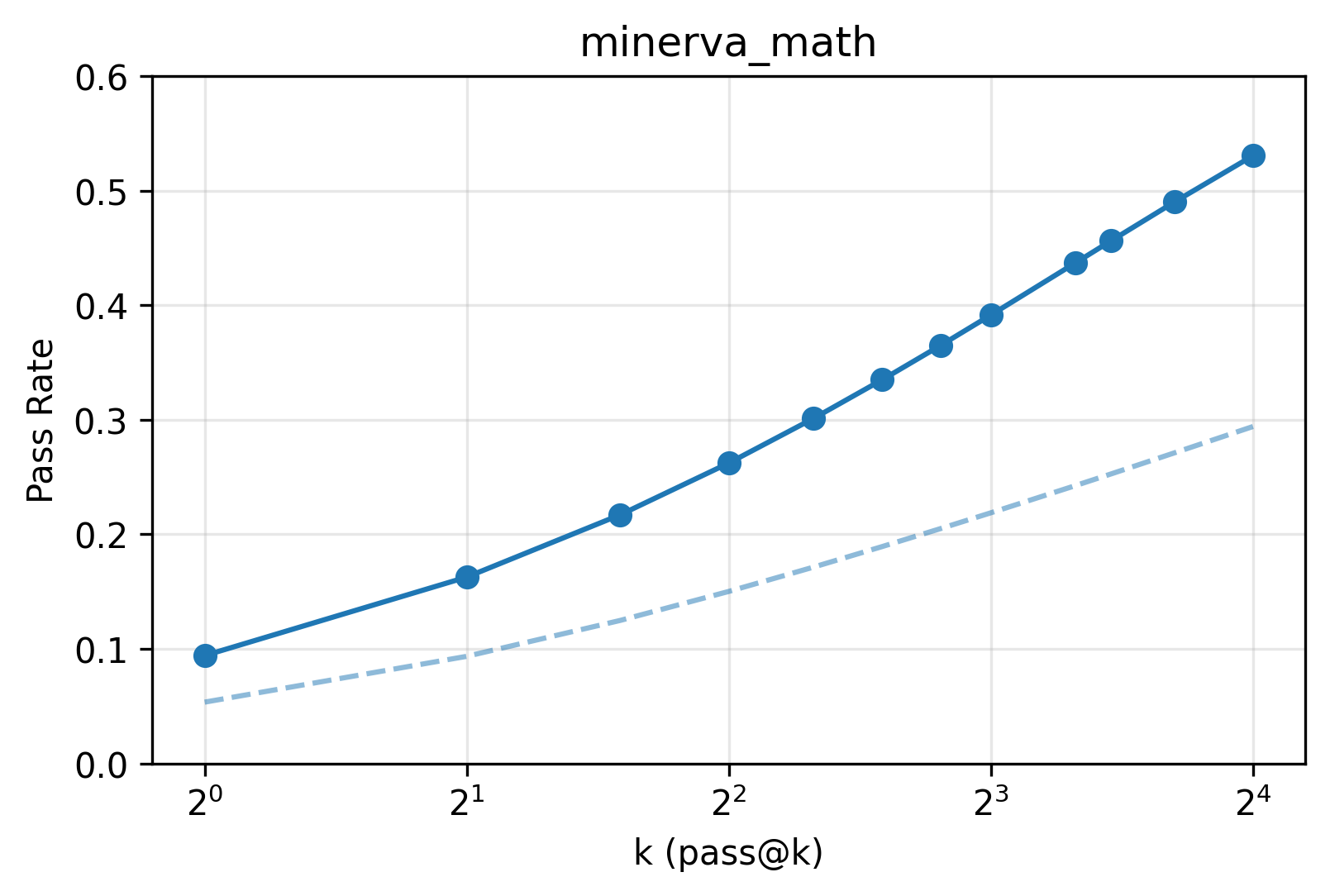} \\
\includegraphics[width=\plotwidthllama]{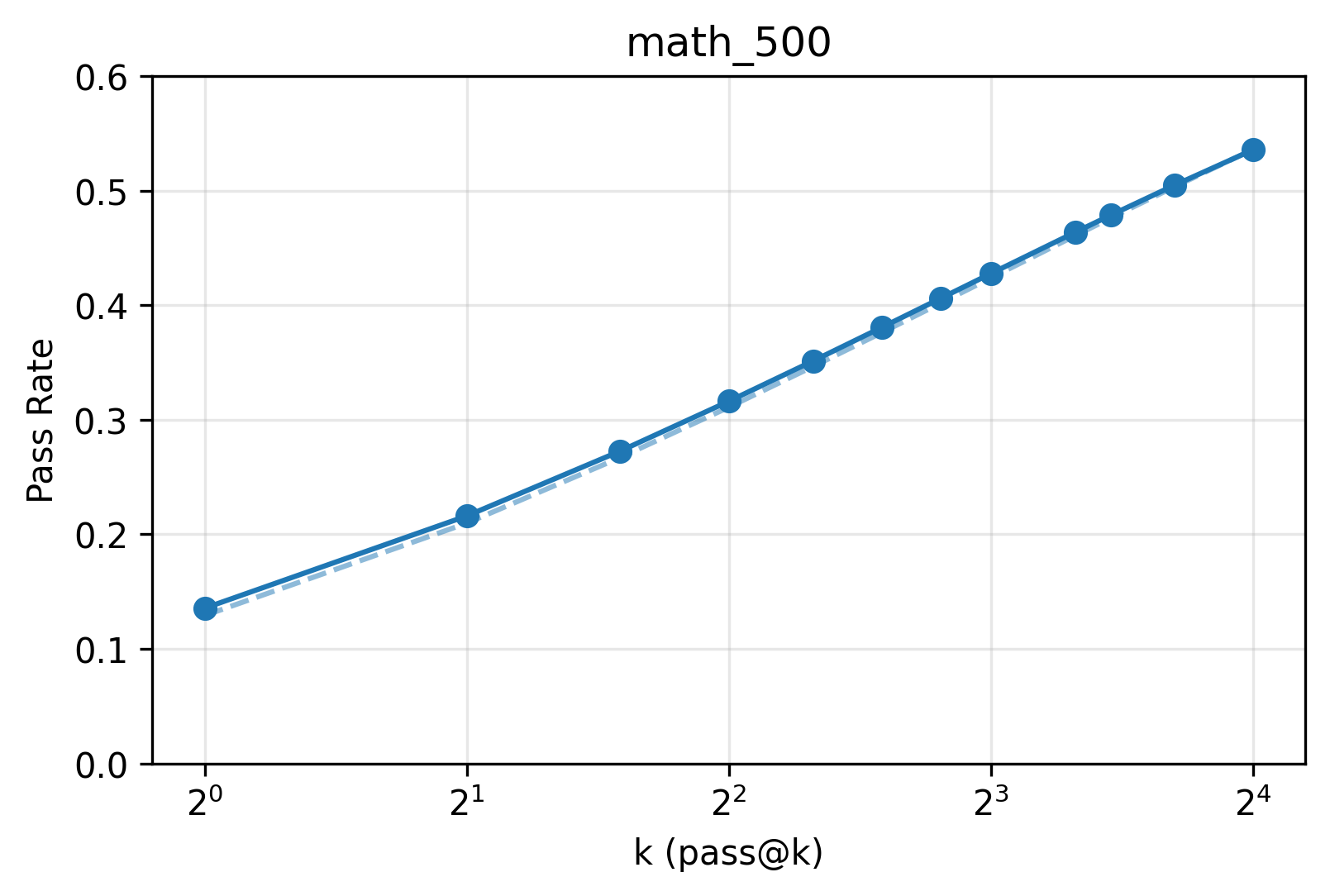} &
\includegraphics[width=\plotwidthllama]{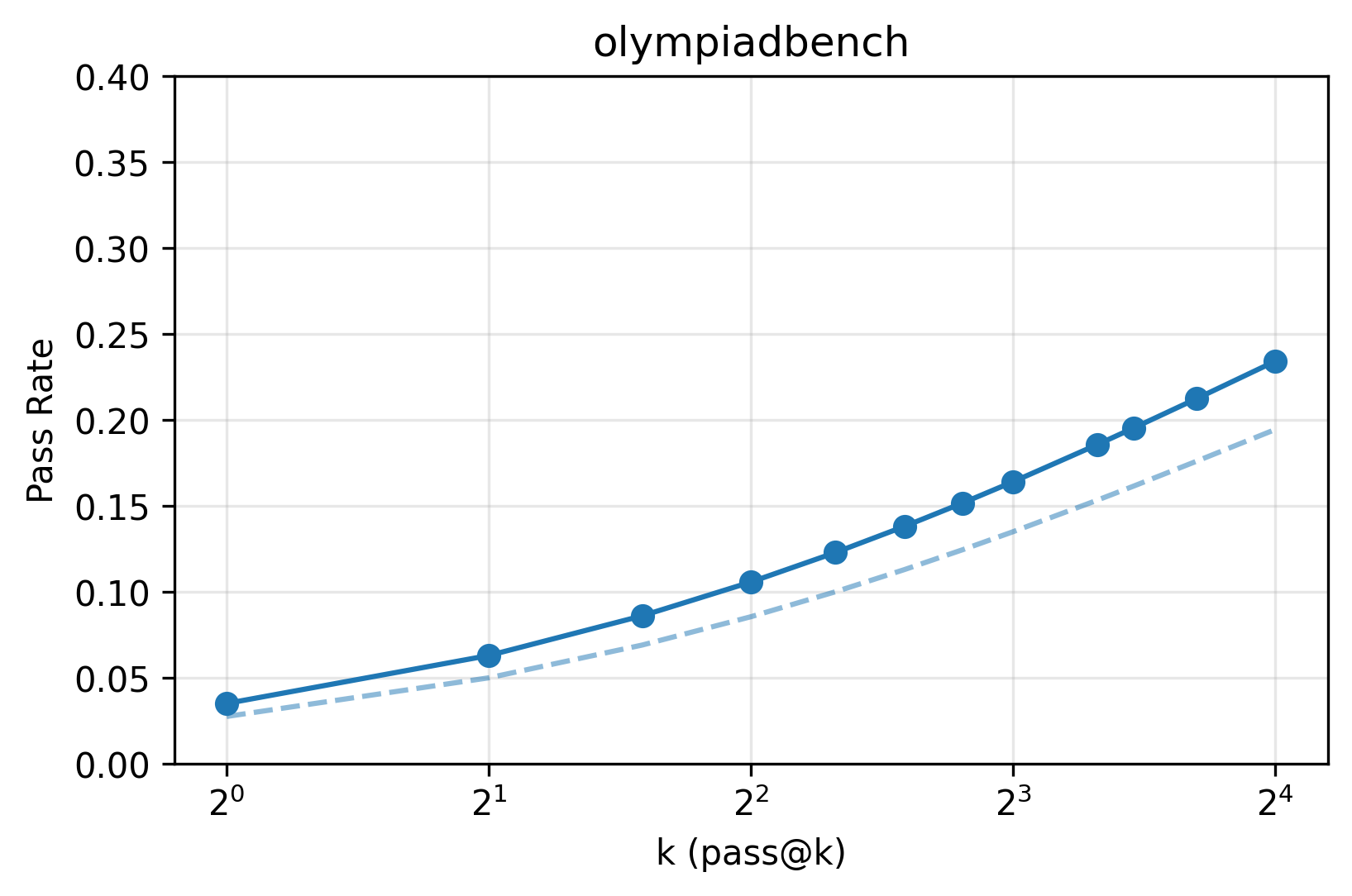} \\
\end{tabular}
\caption{Pass@k evaluation results comparing Lighteval package \cite{lighteval} baseline evaluation (dashed line) with our LLM-as-a-judge evaluation approach (solid line) on the Llama3.1-8B model \cite{grattafiori2024llama} across four mathematical reasoning datasets: GSM8K, Minerva, Math500, and Olympiad.}
\label{fig:lighteval_llama_results}
\end{figure*}

\begin{figure*}[t]
\centering
\setlength{\tabcolsep}{2pt}   %
\newcommand{\plotwidthapp}{0.32\textwidth}
\newcommand{\raisetxtapp}{0.6cm}
\begin{tabular}{ccc}
\small Qwen2.5-7B &\small Qwen2.5-14B & \small Qwen2.5-32B \\
\includegraphics[width=\plotwidthapp]{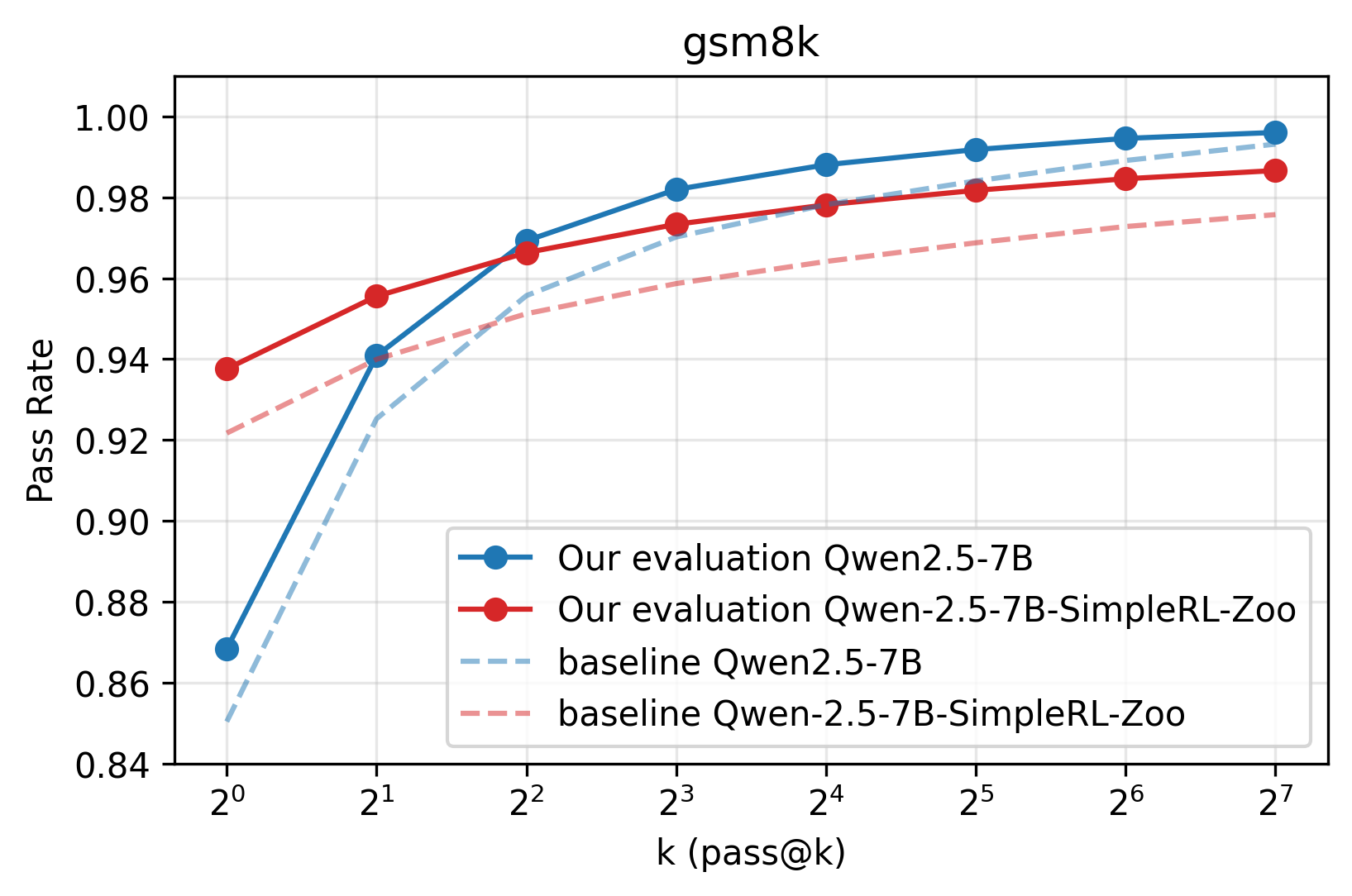} &
\includegraphics[width=\plotwidthapp]{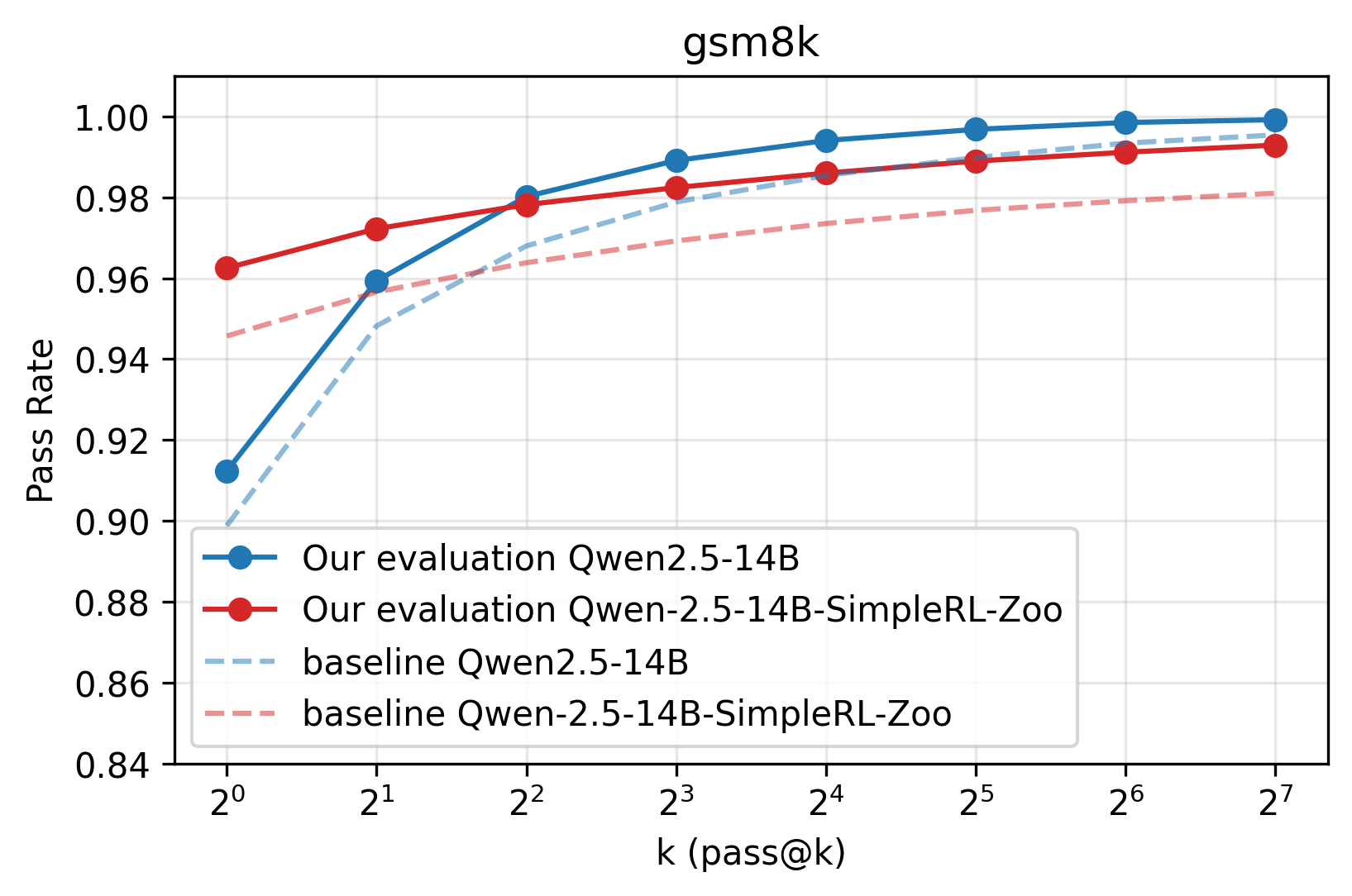} &
\includegraphics[width=\plotwidthapp]{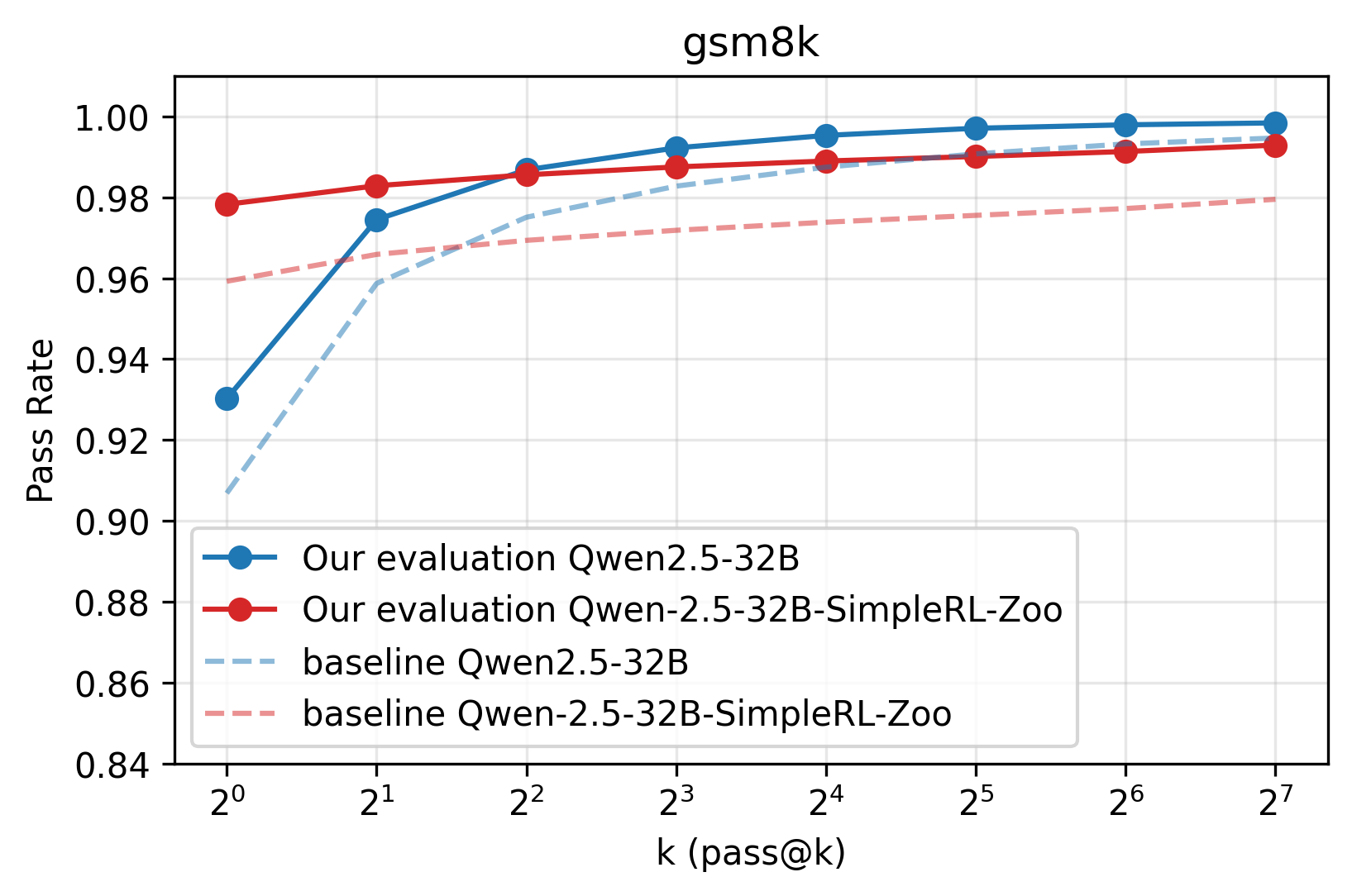} \\
\includegraphics[width=\plotwidthapp]{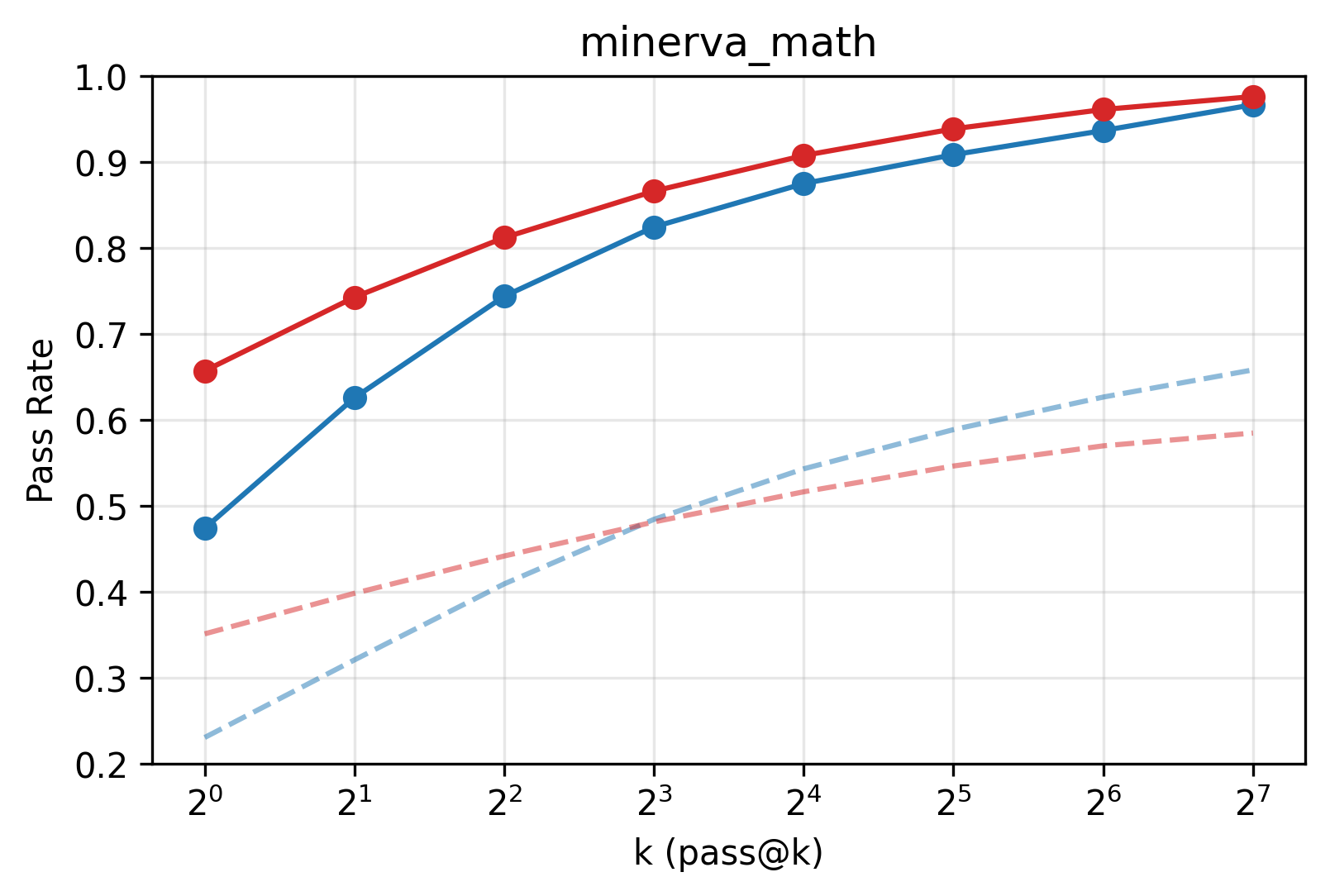} &
\includegraphics[width=\plotwidthapp]{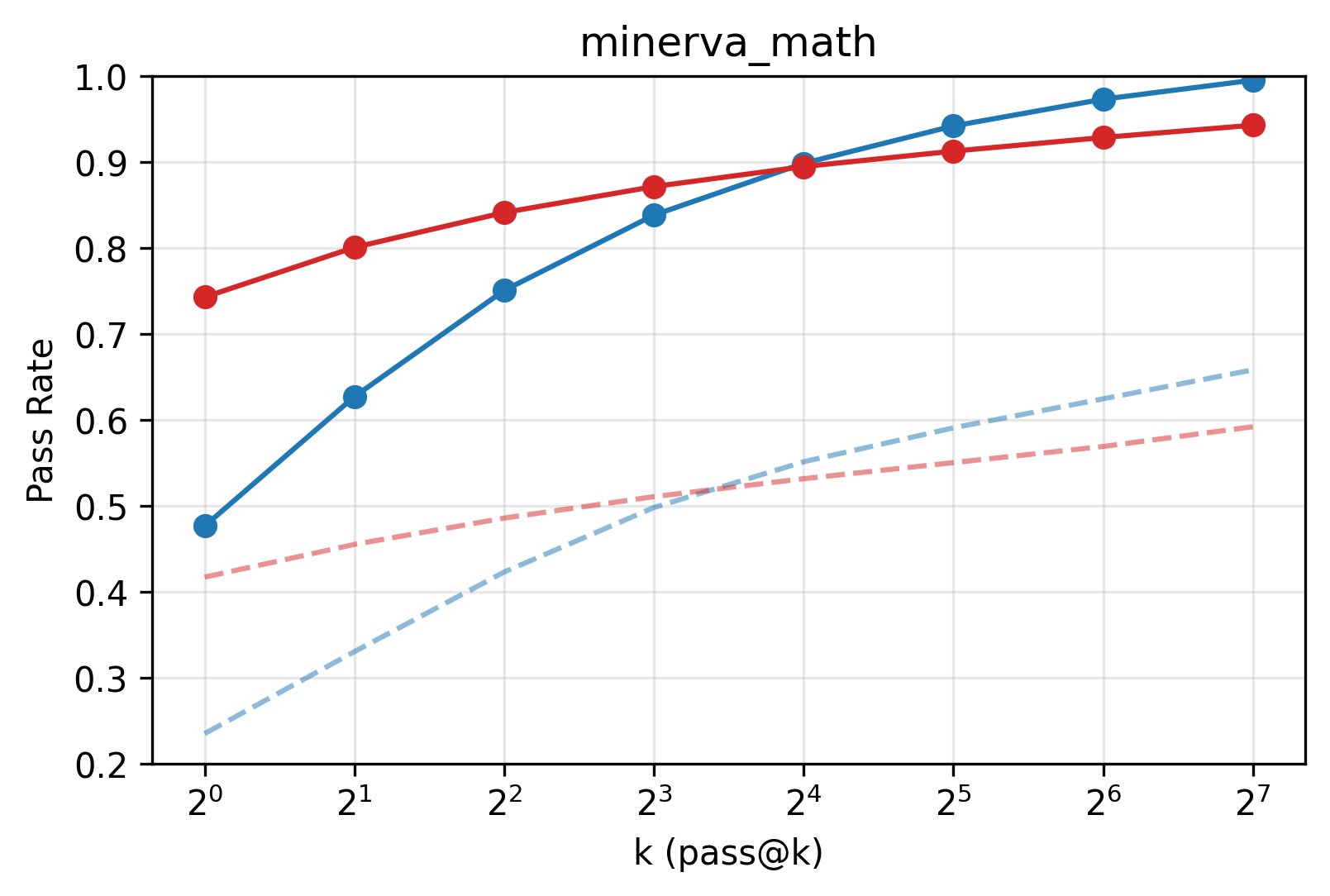} &
\includegraphics[width=\plotwidthapp]{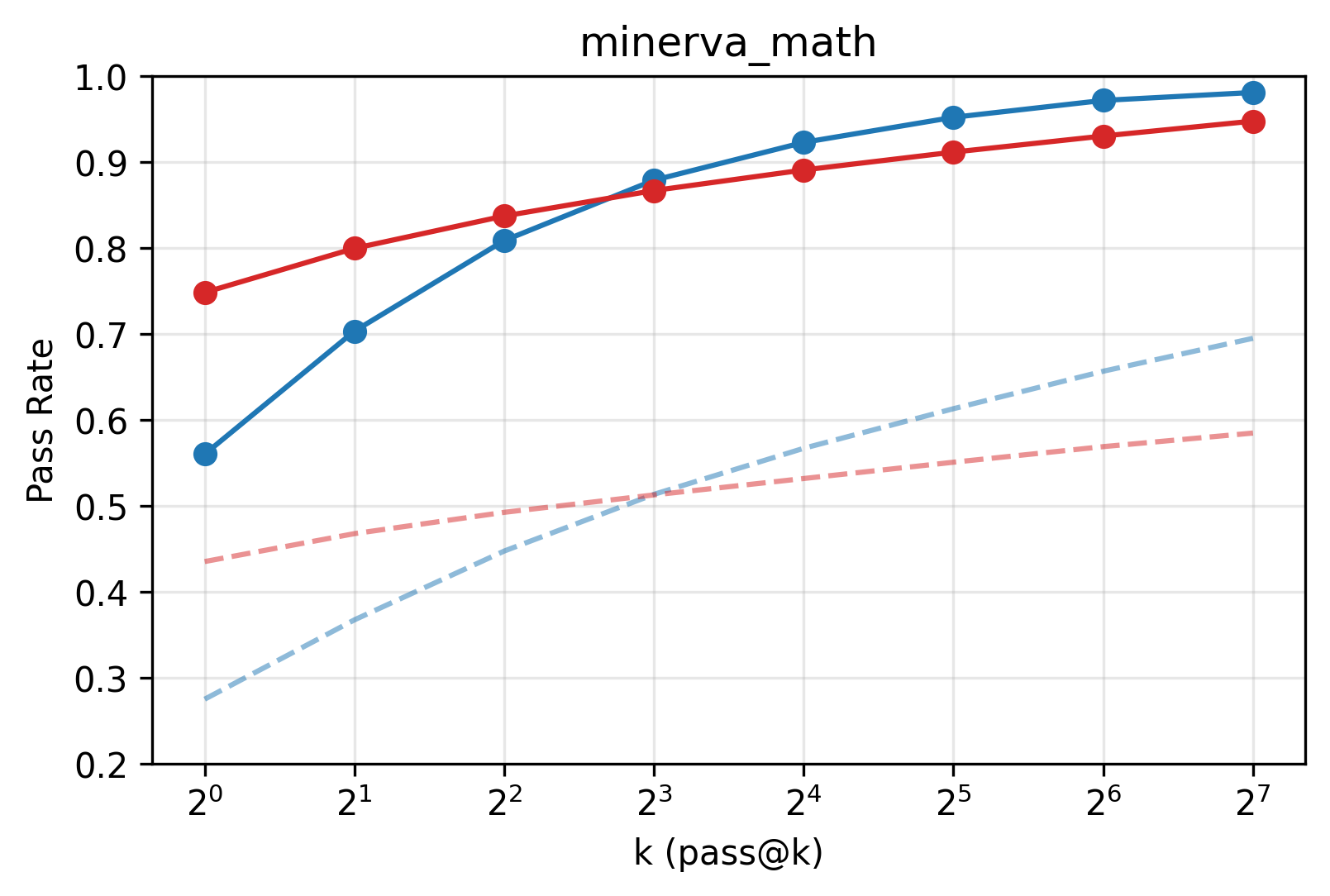} \\
\includegraphics[width=\plotwidthapp]{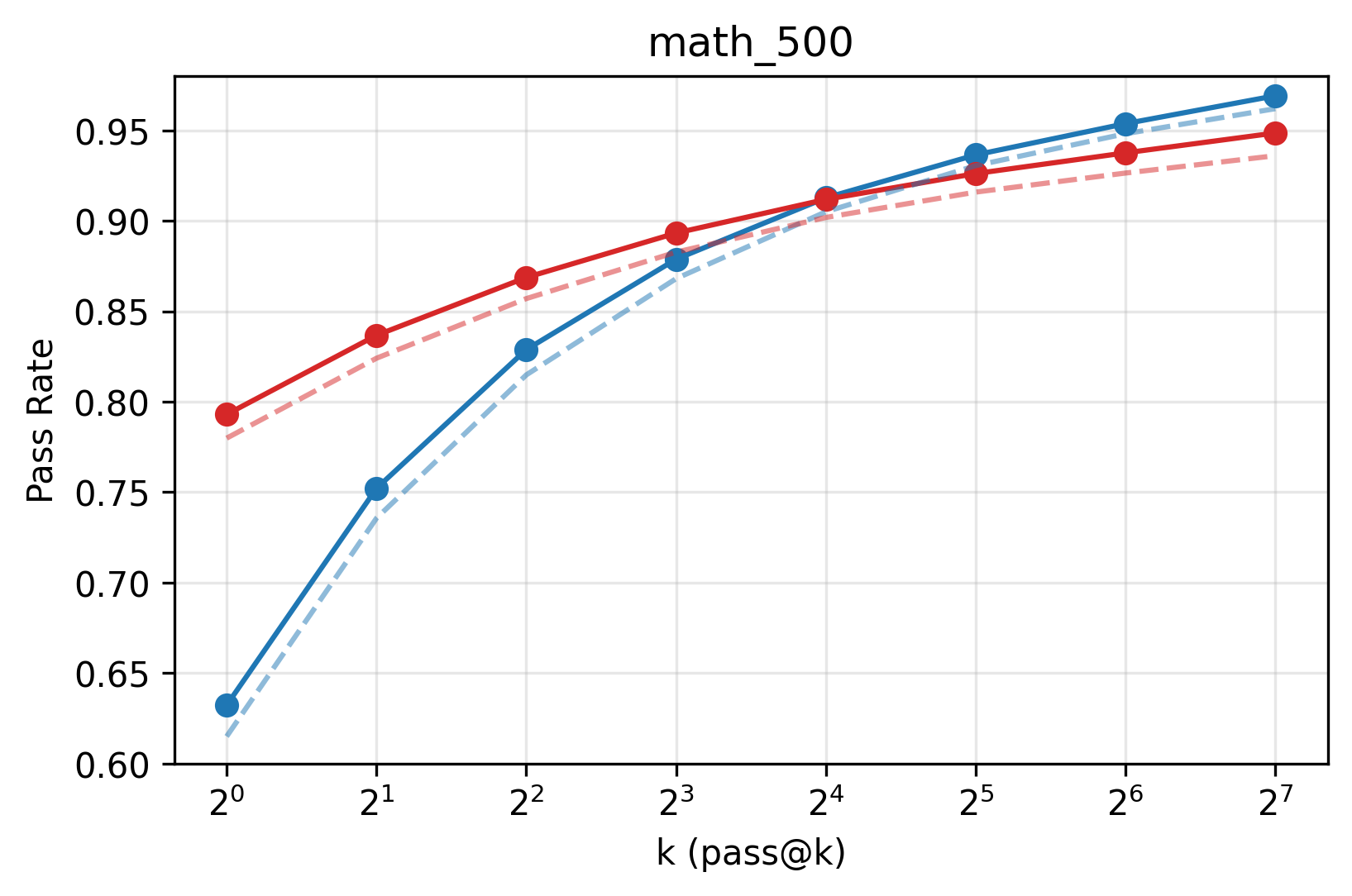} &
\includegraphics[width=\plotwidthapp]{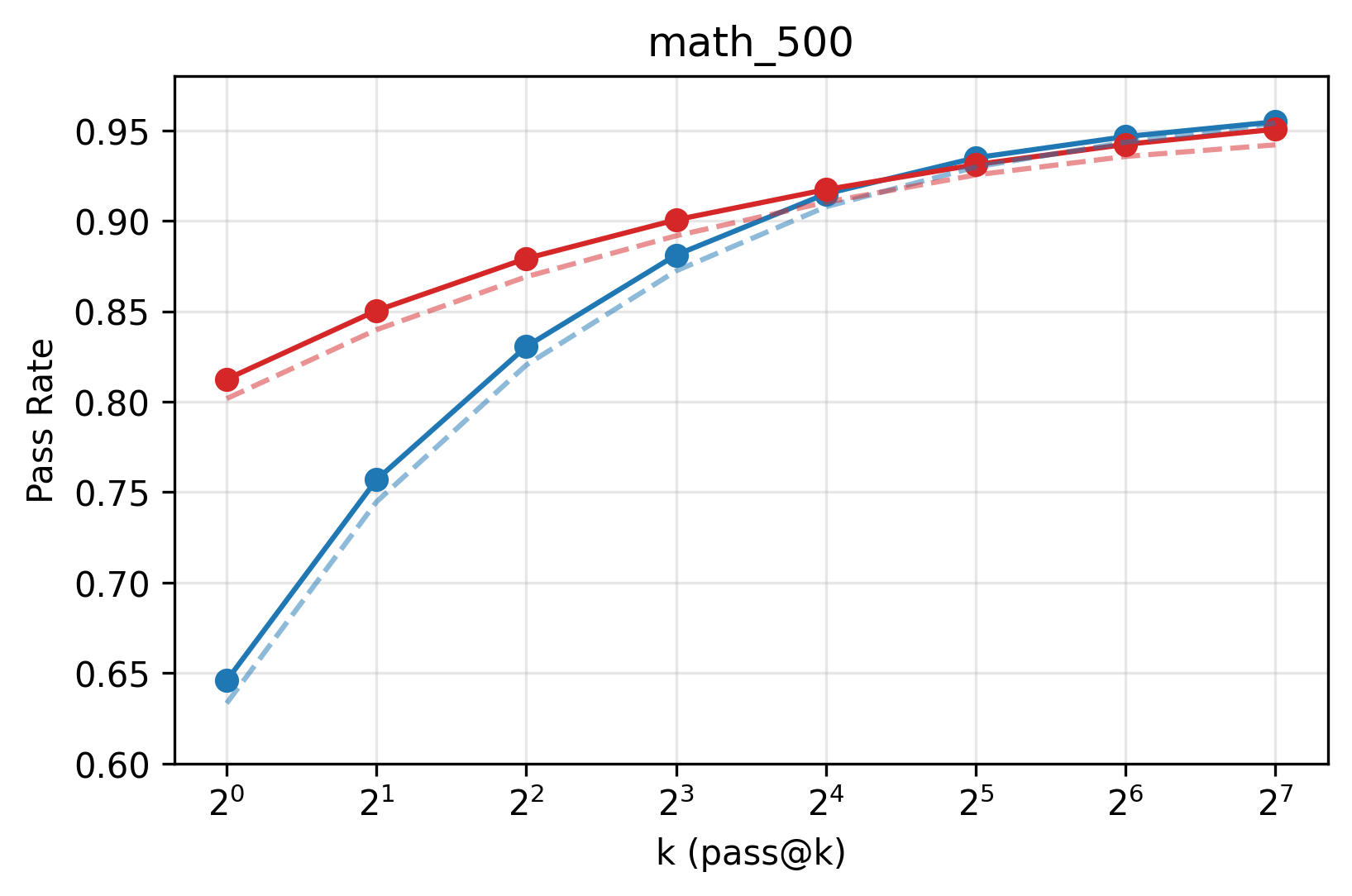} &
\includegraphics[width=\plotwidthapp]{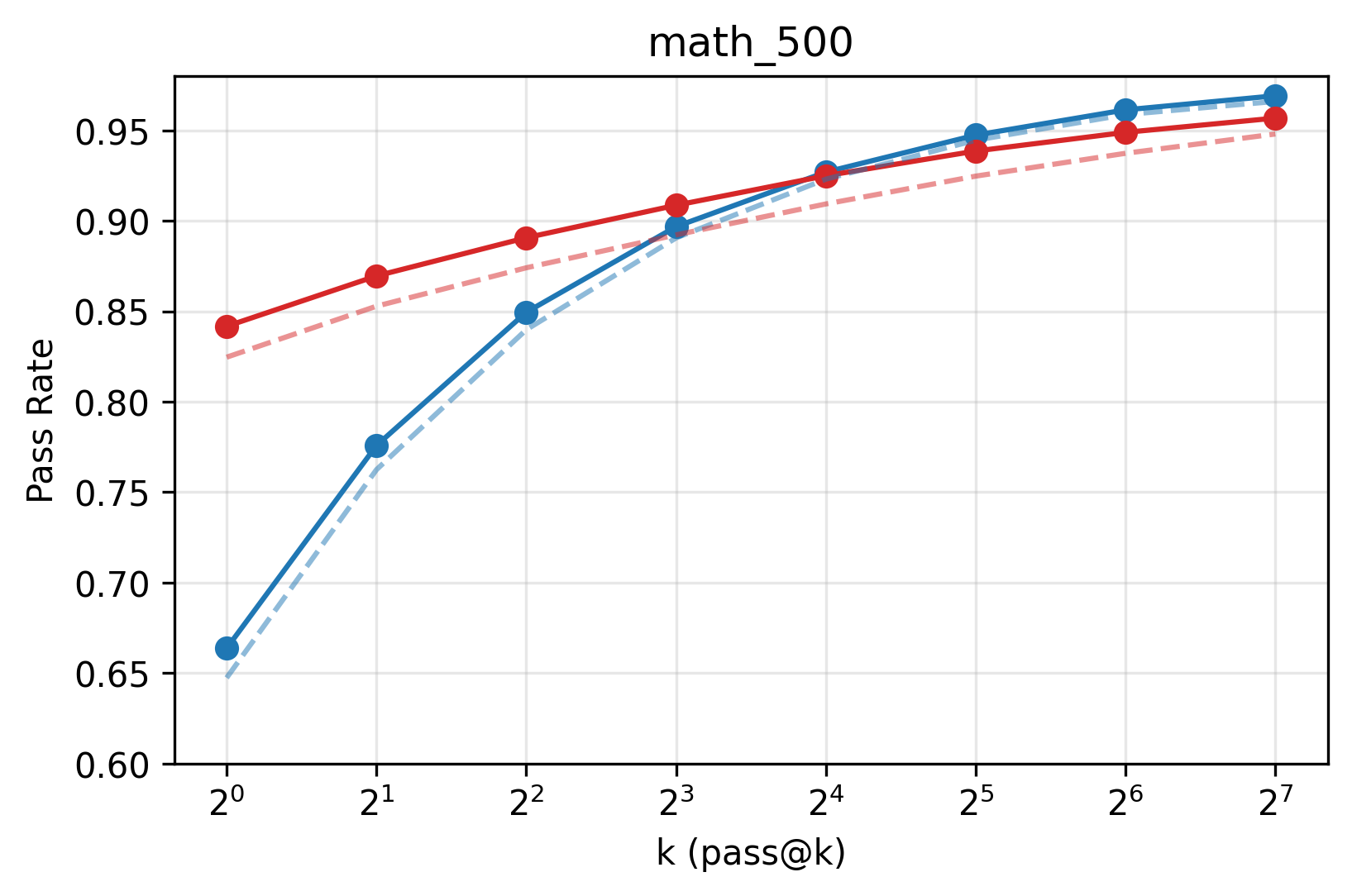} \\
\includegraphics[width=\plotwidthapp]{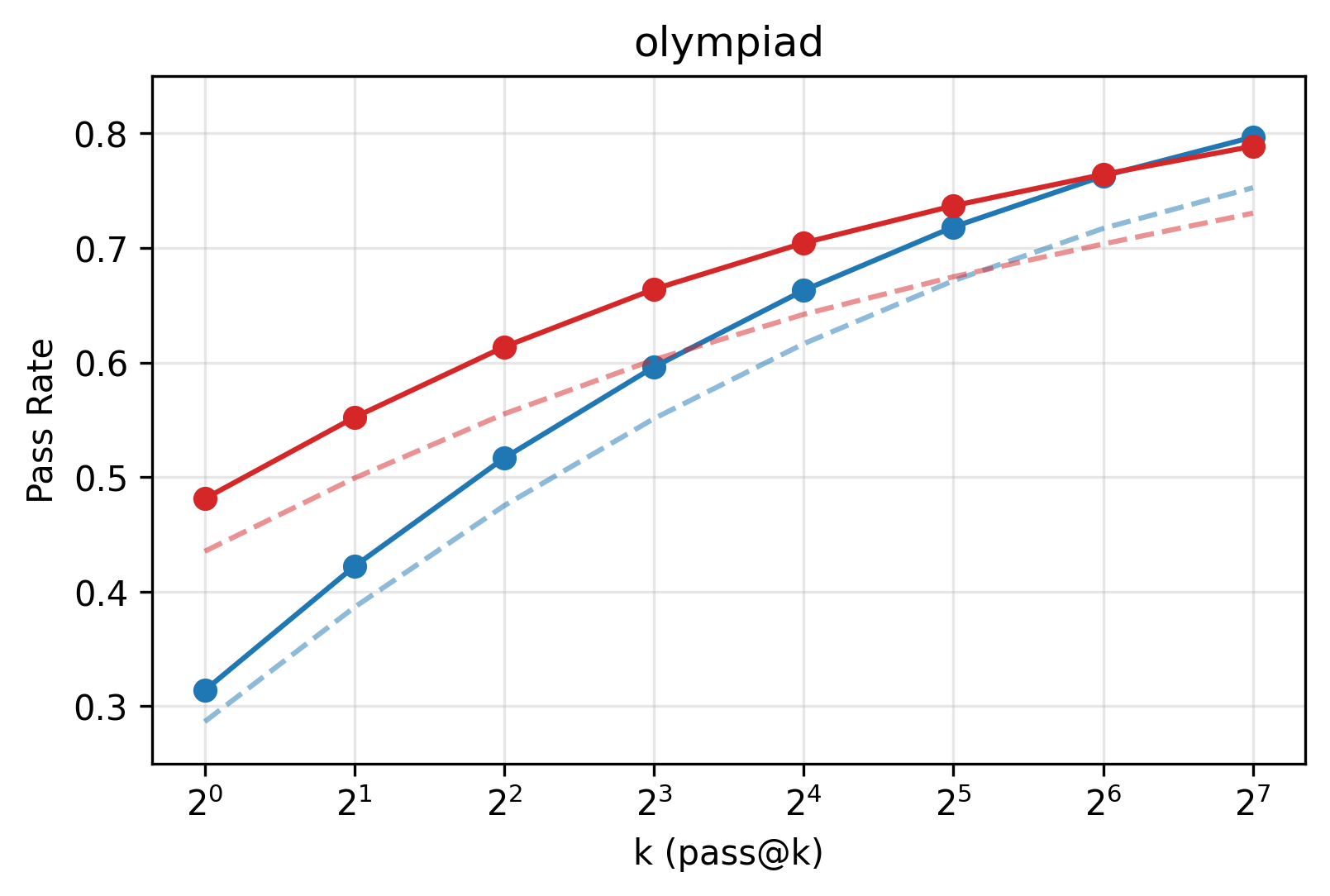} &
\includegraphics[width=\plotwidthapp]{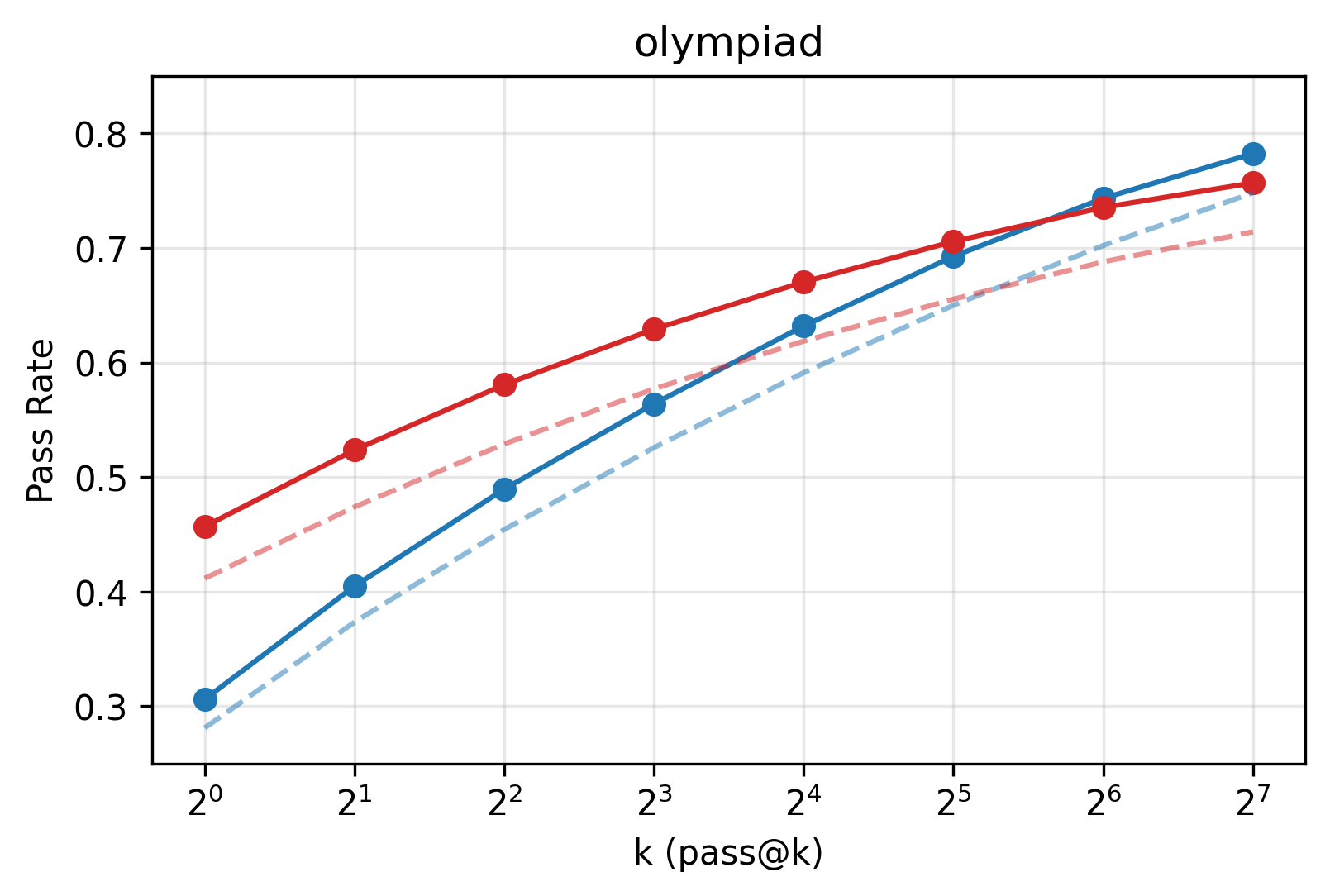} &
\includegraphics[width=\plotwidthapp]{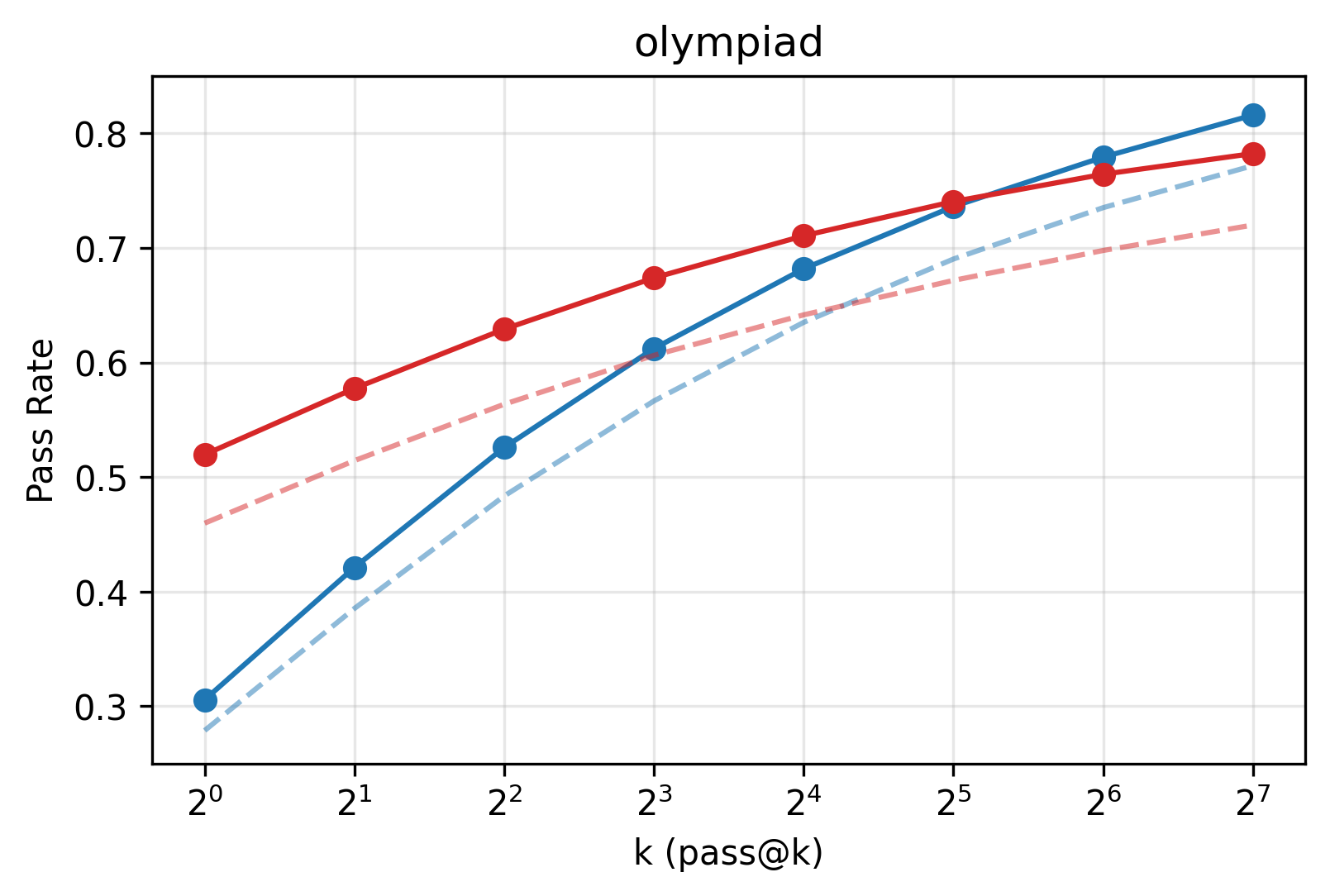} \\
\end{tabular}
\caption{Pass@k evaluation results for Qwen2.5-7B, 14B, and 32B models comparing baseline evaluation method (dashed line) with our LLM-as-a-judge evaluation approach (solid line) on GSM8K, Minerva, Math500, and Olympiad datasets (top to bottom rows). Our approach consistently outperforms baseline evaluation across all three model sizes and all datasets. We evaluated the models up to $k=128$, following prior popular work \cite{yue2025does}, which uses the SimpleRL evaluation framework, and we reproduce their reported results.}
\label{fig:passatk_results_14b_32b}
\end{figure*}

\begin{figure*}[t]
\centering
\newcommand{\plotwidthmath}{0.4\textwidth}
\begin{tabular}{cc}
\small SimpleRL Evaluation & \small Lighteval Evaluation \\
\includegraphics[width=\plotwidthmath]{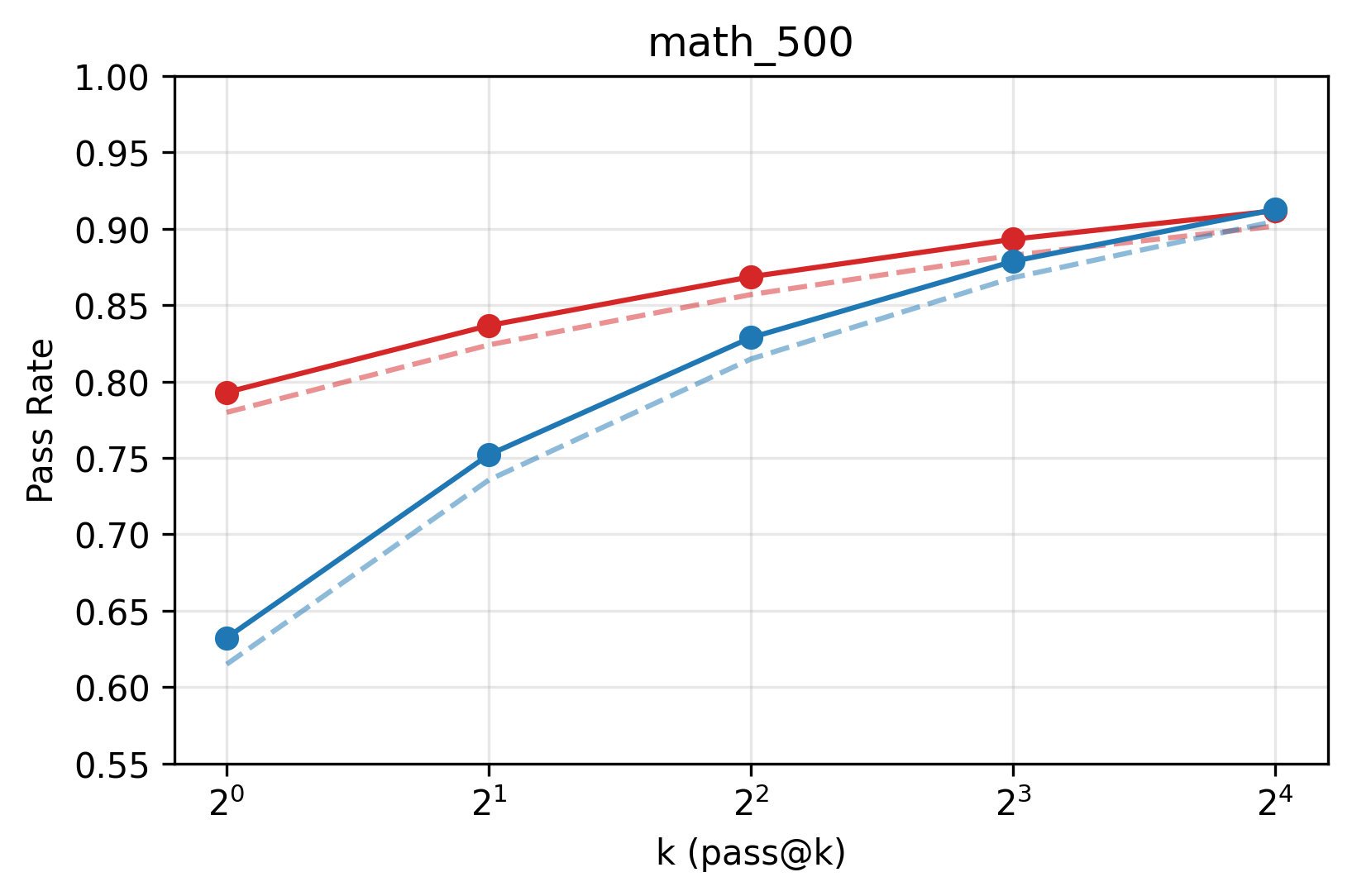} &
\includegraphics[width=\plotwidthmath]{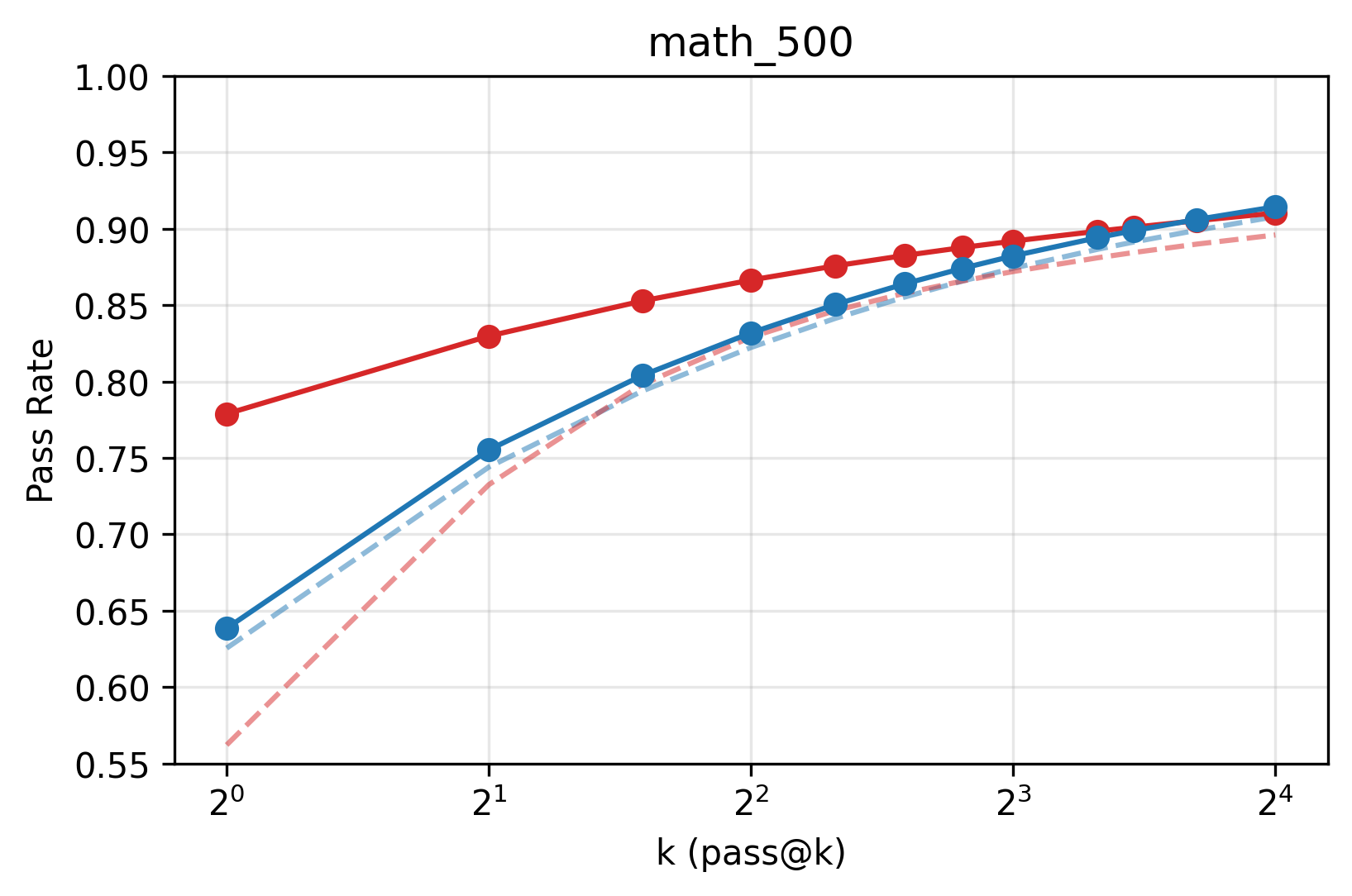} \\
\includegraphics[width=\plotwidthmath]{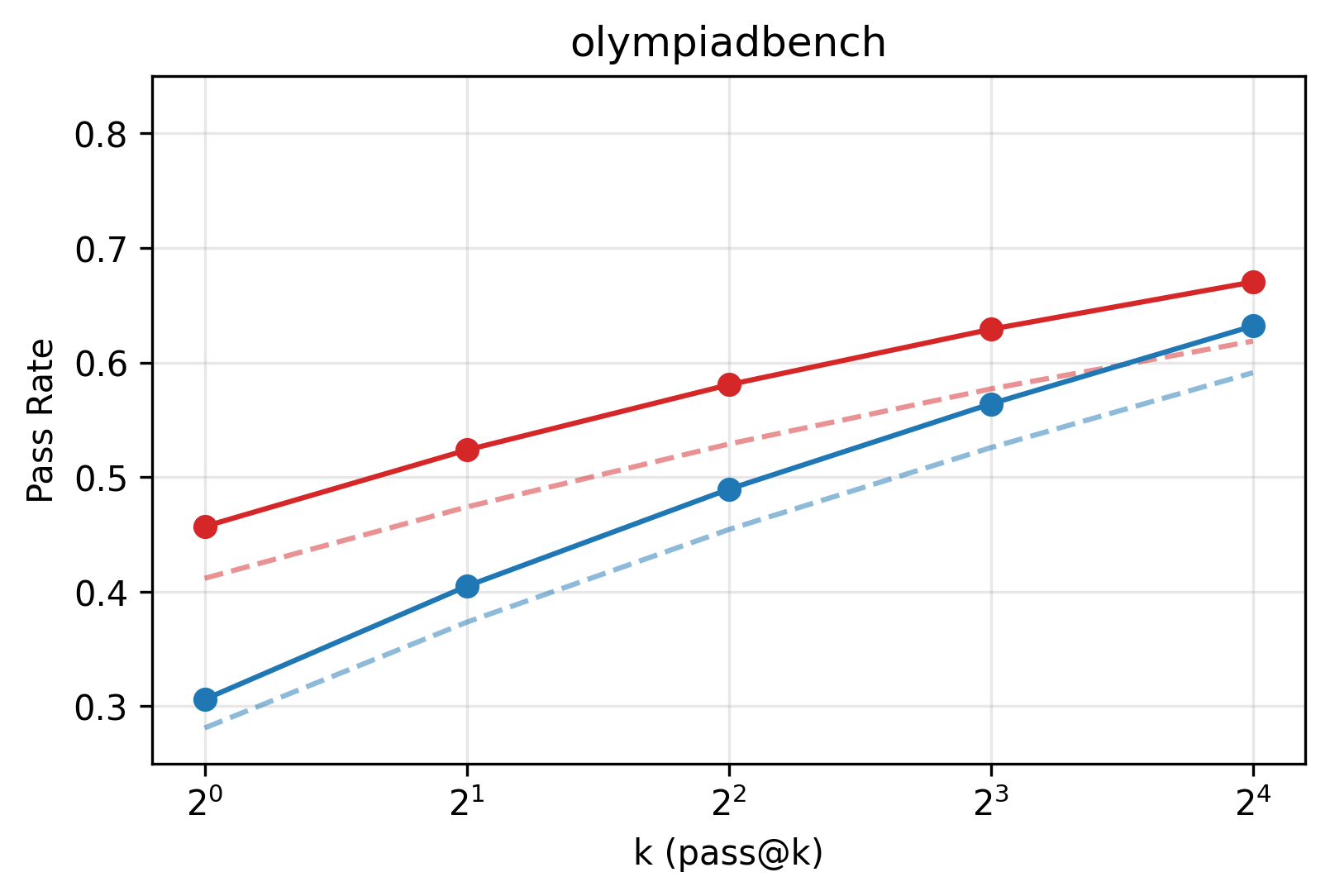} &
\includegraphics[width=\plotwidthmath]{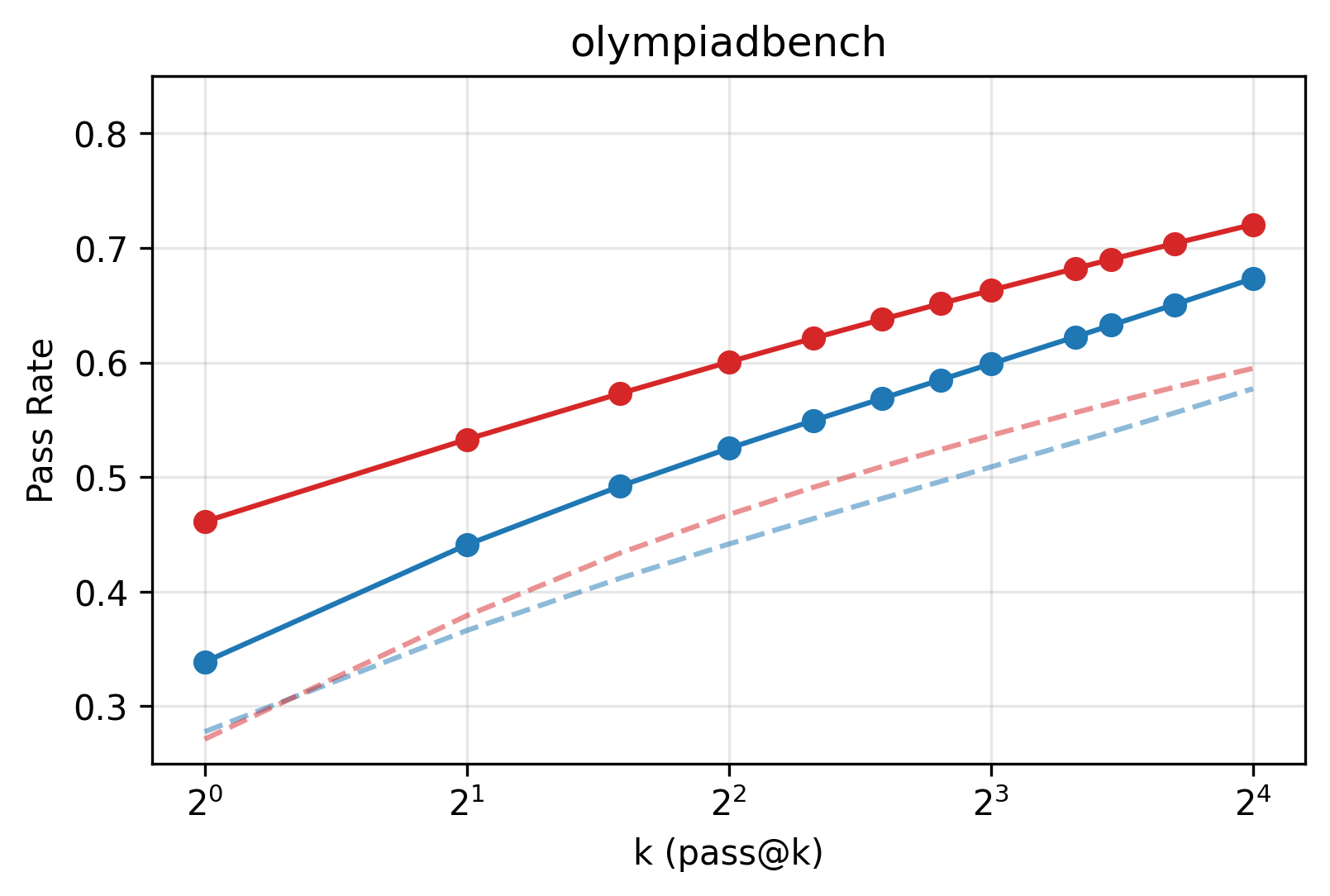} \\
\end{tabular}
\caption{Pass@k evaluation results comparing baseline evaluation methods (dashed line) with our LLM-as-a-judge evaluation approach (solid line) on Qwen2.5-7B model. The left column displays results obtained using the SimpleRL evaluation framework, while the right column presents the results from the Lighteval evaluation framework. Two datasets are presented: Math500 (top) and Olympiad (bottom). Our approach presents consistent evaluation across both frameworks and outperforms baseline evaluation.
}
\label{fig:passatk_7b_results_appendix}
\end{figure*}

\section{Additional Pass@k Evaluations}

We provide additional pass@k evaluation results to complement the findings presented in the main paper.

\Cref{fig:lighteval_llama_results} presents the evaluation results for the Llama3.1-8B \cite{grattafiori2024llama} model across four mathematical reasoning datasets using the Lighteval framework \cite{lighteval}. Our LLM-as-a-judge approach (solid line) demonstrates similar results to the Qwen2.5 model family. \Cref{tab:qexample4} presents examples of our approach improvement compared to the Lighteval symbolic evaluator for Llama3.1-8B model.

\Cref{fig:passatk_results_14b_32b} extends our analysis to larger model sizes, showing pass@k results for Qwen2.5-7B, 14B, and 32B models \cite{qwen2024qwen2} on GSM8K, Minerva, Math500, and Olympiad datasets. \Cref{fig:passatk_7b_results_appendix} presents pass@k results for the Qwen2.5-7B model on Math500 and Olympiad datasets, comparing evaluations from both the SimpleRL and Lighteval frameworks. Similar to the GSM8K and Minerva results presented in the main paper (\Cref{fig:passatk_7b_results1}), our LLM-as-a-judge approach provides consistent evaluation regardless of the framework used, while the baseline methods exhibit notable discrepancies between frameworks. This consistency further validates the robustness of our approach across different evaluation pipelines.

\section{Meta-Evaluation Classification Examples}

\Cref{tab:classification} presents representative examples from our meta-evaluation dataset, illustrating the classification outcomes of our LLM-as-a-judge approach. For each question, we show the GT answer from the dataset alongside various model predictions and their corresponding evaluation results categorized as true positives, false positives, and false negatives (true negative was omitted).

These examples highlight the inherent challenge in mathematical answer verification: while the dataset provides a single fixed GT answer, multiple predictions may be mathematically correct due to rounding differences, precision variations, or equivalent representations. For instance, in the first example (GT: 50), predictions ranging from 48.48 to 48.5 are correctly identified as true positives by our approach, as the more precise calculation yields approximately 48.48 microns rather than the rounded GT value of 50.

The table also reveals the boundaries of acceptable tolerance in our evaluation. In the second example, predictions of 6.61 and 6.62 are accepted as correct, while 6.59 and 6.63 are classified as false positives, demonstrating the LLM judge's ability to apply reasonable precision thresholds. The third example shows a rare false negative case, where the inclusion of units (``m'') in an otherwise correct answer led to misclassification—an edge case that informs potential improvements to our framework.

These examples demonstrate both the capabilities and limitations of our LLM-based evaluation approach and justify the design choices validated through our ablation study in \Cref{tab:ablations}.

\begin{table*}[bt]
\centering
\caption{Example questions and predictions from our meta-evaluation dataset, and the results of our LLM-based binary verification. In our manually labeled dataset, various predictions are considered correct (due to different formatting and accuracy levels), in contrast to the dataset GT, which provides only a single fixed answer. }
\label{tab:classification}
\setlength{\tabcolsep}{4pt}
\begin{tabular}{p{4.5cm}p{10cm}}
\hline
\multicolumn{2}{p{14cm}}{\small \textbf{Question:} A large ground-based telescope has an effective focal length of 10 meters. Two astronomical objects are separated by 1 arc second in the sky. How far apart will the two corresponding images be in the focal plane, in microns? \; \textcolor{navyblue}{GT: 50}}  \\
\hline
LLM-as-a-Judge Evaluation & Prediction Values \\
\hline
\small True Positive & \small 48.48, \; 48.48136811, \; 48.48137, \; 48.481, \; 48.5 \\
\small False Positive & \small 48, \; 48.6, \; 48, \; 49.348, \; 48.504, \; 49.348, \; 48, \; 48.56, \; 48 \\
\small False Negative & \small - \\
\hline\hline
\multicolumn{2}{p{14cm}}{\small \textbf{Question:} Determine the minimum potential in V (to 2 decimal places) that must be applied to an $\alpha$-particle so that on interaction with a hydrogen atom, a ground state electron will be excited to $n$ $=6$. \; \textcolor{navyblue}{GT: 6.62}} \\
\hline
LLM-as-a-Judge Evaluation & Prediction Values \\
\hline
\small True Positive & \small 6.61, \; 6.62 \\
\small False Positive & \small 6.59, \; 6.63 \\
\small False Negative & \small - \\
\hline\hline
\multicolumn{2}{p{14cm}}{\small \textbf{Question:} Determine the atomic (metallic) radius of Mo in meters. Do not give the value listed in the periodic table; calculate it from the fact that Mo's atomic weight is $=95.94 \mathrm{~g} /$ mole and $\rho=10.2 \mathrm{~g} / \mathrm{cm}^{3}$. Please format your answer as $n \times 10^x$ where $n$ is to 2 decimal places. \; \textcolor{navyblue}{GT: 1.39e-10}} \\
\hline
LLM-as-a-Judge Evaluation & Prediction Values \\
\hline
\small True Positive & \small 1.36 $\times$ 10$^{-10}$, \; 1.37 $\times$ 10$^{-10}$, \; 1.37 * 10$^{-10}$ \\
\small False Positive & \small - \\
\small False Negative & \small 1.36 $\times$ 10$^{-10}$ m \\
\hline
\end{tabular}
\end{table*}

\begin{table*}[!tb]
\centering
\caption{An example question illustrating the difference between symbolic and LLM-based evaluation approaches. Our approach generalizes to various answer representations. The symbolic evaluation accepts only the result that matches the GT answer explicitly.}
\label{tab:qexample2}
\setlength{\tabcolsep}{4pt}
\begin{tabular}{p{8cm} p{2cm} p{2cm}}
\hline
\multicolumn{3}{p{14cm}}{\small \textbf{Question:} "Britany records 18 4-minute TikTok videos each week. She spends 2 hours a week writing amateur songs to sing on TikTok, and 15 minutes six days a week doing her makeup before filming herself for TikTok. How much time does Britany spend on TikTok in a month with four weeks?" \; \textcolor{navyblue}{GT: 1128} \; {\footnotesize (GSM8K)} } \\
\hline
Answer & Symbolic & LLM \small (Ours) \\
\hline
\formatcorrecttext & & \\
\small\verb|1128| & \Large \textcolor{darkgreen}{\cmark} & \Large \textcolor{darkgreen}{\cmark} \\
\small\verb|18.8| & \Large \textcolor{red}{\xmark} & \Large \textcolor{darkgreen}{\cmark} \\
\small\verb|18 \text{ hours and } 48 \text{ minutes}| & \Large \textcolor{red}{\xmark} & \Large \textcolor{darkgreen}{\cmark} \\
\small\verb|18 \text{ hours } 48 \text{ minutes}| & \Large \textcolor{red}{\xmark} & \Large \textcolor{darkgreen}{\cmark} \\
\small\verb|18 hours and 48 minutes| & \Large \textcolor{red}{\xmark} & \Large \textcolor{darkgreen}{\cmark} \\
\formatincorrecttext & & \\
\small\verb|14| & \Large \textcolor{red}{\xmark} & \Large \textcolor{red}{\xmark} \\
\small\verb|48| & \Large \textcolor{red}{\xmark} & \Large \textcolor{red}{\xmark} \\
\small\verb|858| & \Large \textcolor{red}{\xmark} & \Large \textcolor{red}{\xmark} \\
\hline
\end{tabular}
\end{table*}

\section{Additional Evaluation Examples}
\label{sec:appendix_examples}

We provide additional examples demonstrating the differences between symbolic and LLM-based evaluation across various scenarios and datasets.

\subsection{Diverse Answer Representations}

As detailed in the main paper and presented in \Cref{tab:examples1}, symbolic evaluation is sensitive to answer representations and fails to recognize mathematically equivalent formats while our approach is more accurate and robust for such formatting differences. We provide additional examples demonstrating this limitation in \Cref{tab:qexample2} for unit conversions and textual representations, \Cref{tab:qexample3} for comparisons with the Lighteval framework~\cite{lighteval}, and \Cref{tab:qexample4} for Llama3.1-8B model \cite{grattafiori2024llama} outputs evaluated with Lighteval. These examples consistently show that our LLM-based approach generalizes across diverse answer formats where symbolic evaluation fails.

\subsection{False Positive Cases in Symbolic Evaluation}

Beyond rejecting correct answers, symbolic evaluation can also incorrectly accept wrong answers, even though it is less common. \Cref{tab:minerva_example} illustrates two such cases. In the first example from Minerva, the question explicitly requires the answer to be formatted as ``+n'' or ``-n''. The symbolic evaluator incorrectly accepts ``2'' as correct, ignoring the formatting requirement that specifies a sign prefix. Our LLM-based approach correctly rejects this answer since it does not adhere to the requested format. The second example involves interval notation where the GT answer is $(-\infty, 0]$. The symbolic evaluator incorrectly accepts $(-\infty, 0)$ as equivalent, failing to distinguish between the closed bracket (including 0) and the open parenthesis (excluding 0). This mathematically significant difference is correctly identified by our approach, which rejects the incorrect interval notation.

\subsection{Ambiguous Questions and Dataset Limitations}

\Cref{tab:ambiguous_example} presents cases where reference answers may be inaccurate due to ambiguous problem statements in the dataset. In the first example, the question asks about pizzas for ``20 friends,'' but it is unclear whether John himself should be included in the count. The GT answer assumes John is excluded (10 pizzas for 20 people), while an equally valid interpretation includes John (11 pizzas for 21 people).

The second example exhibits a similar ambiguity: the question asks how many vines ``he'' (Steve) needs, but the GT answer includes both Steve's and his girlfriend's consumption. A literal interpretation of the question would yield 14 vines (for Steve's 42 tomatoes per week), not the provided answer of 21.
These cases highlight the importance of our answer validation stage (\Cref{sec:methods:s2}), which allows the LLM to identify potentially incorrect or ambiguous GT answers.

\section{Computational Resources}
\label{sec:computational_resources}

\paragraph{Model Inference.}
We evaluated models from the Qwen2.5 family with 7B, 14B, and 32B parameters, as well as Llama3.1-8B (~8B parameters), including their RLVR fine-tuned variants. Model inference for generating predictions was conducted on NVIDIA A100 GPUs (8 GPUs).

\paragraph{LLM-as-a-Judge Evaluation.}
For the evaluation pipeline, we used Claude Sonnet 4 as the primary LLM judge, accessed via API. Additional validation experiments were conducted with Mistral and Llama models, also via API calls. The number of API calls scales with the dataset sizes (GSM8K, Math500, Olympiad, and Minerva), the number of sampled predictions per question for pass@k computation, and the evaluation parameters described in \Cref{sec:methods:s3} (group size $n_g$ and number of verifications $n_{verif}$ for majority voting).

\section{Dataset Validation and Cleaning}
\label{sec:appendix_dataset_validation}

We acknowledge that erroneous questions and annotations in existing benchmarks can harm evaluation correctness, as illustrated in \Cref{tab:unclear_questions} and \Cref{tab:ambiguous_example}. However, we note that dataset validation and cleaning is not the primary contribution of our work; rather, our focus is on the LLM-as-a-judge evaluation framework and overcoming symbolic evaluation limitations.

To address this, we apply an LLM-based dataset validation procedure as described in \Cref{sec:methods:s2}. Alternative approaches for dataset cleaning, such as those proposed by \citet{shen2025let,zheng2025processbench}, can also be applied. This filtering improves evaluation accuracy by excluding potentially erroneous annotations, as observed in our experiments. Filtered questions can be manually reviewed by domain experts and reintroduced to the dataset if appropriate.
We emphasize that the numerical improvement of our method over symbolic baselines is not solely an indicator of better evaluation. To validate our approach, we present concrete failure case examples and demonstrate consistency across different evaluation frameworks and models.

\begin{table}[t]
\centering
\caption{Examples of \emphtxt{false positive symbolic evaluation}. The symbolic evaluation incorrectly accepts inaccurate answers.
In the first example, the symbolic evaluation fails to enforce formatting requirements, and in the second, it accepts mathematically incorrect interval notation. Our LLM-based approach correctly identifies such errors.}
\label{tab:minerva_example}
\setlength{\tabcolsep}{4pt}
\begin{tabular}{p{3cm} p{1.5cm} p{1.9cm}}
\hline
\multicolumn{3}{p{7.2cm}}{\small \textbf{Question:} "What is the net charge of arginine in a solution of pH 1.0? Please format your answer as +n or -n." \; \textcolor{navyblue}{GT: +2} \; {\footnotesize (Minerva)}} \\
\hline
Answer & Symbolic & LLM \small (Ours) \\
\hline
\formatcorrecttext & & \\
\small\verb|+2| & \Large \textcolor{darkgreen}{\cmark} & \Large \textcolor{darkgreen}{\cmark} \\
\formatincorrecttext & & \\
\small\verb|2| & \Large \textcolor{darkgreen}{\cmark} & \Large \textcolor{red}{\xmark} \\
\small\verb|+1| & \Large \textcolor{red}{\xmark} & \Large \textcolor{red}{\xmark} \\
\small\verb|+3| & \Large \textcolor{red}{\xmark} & \Large \textcolor{red}{\xmark} \\
\small\verb|-1| & \Large \textcolor{red}{\xmark} & \Large \textcolor{red}{\xmark} \\
\small\verb|1| & \Large \textcolor{red}{\xmark} & \Large \textcolor{red}{\xmark} \\
\small\verb|0| & \Large \textcolor{red}{\xmark} & \Large \textcolor{red}{\xmark} \\
\small\verb|3| & \Large \textcolor{red}{\xmark} & \Large \textcolor{red}{\xmark} \\
\hline\hline
\multicolumn{3}{p{7.2cm}}{\small \textbf{Question:} "What is the range of the function $y=\log_2(\sin x)$ for $0^{\circ}<x<180^{\circ}$?" \; \textcolor{navyblue}{GT: (-$\infty$,0]} \; {\footnotesize (Math500)}} \\
\hline
Answer & Symbolic & LLM \small (Ours) \\
\hline
\formatcorrecttext & & \\
\small\verb|(-\infty,0]| & \Large \textcolor{darkgreen}{\cmark} & \Large \textcolor{darkgreen}{\cmark} \\
\formatincorrecttext & & \\
\small\verb|(-\infty,0)| & \Large \textcolor{darkgreen}{\cmark} & \Large \textcolor{red}{\xmark} \\
\small\verb|0| & \Large \textcolor{red}{\xmark} & \Large \textcolor{red}{\xmark} \\
\small\verb|1000| & \Large \textcolor{red}{\xmark} & \Large \textcolor{red}{\xmark} \\
\small\verb|180| & \Large \textcolor{red}{\xmark} & \Large \textcolor{red}{\xmark} \\
\small\verb|3| & \Large \textcolor{red}{\xmark} & \Large \textcolor{red}{\xmark} \\
\small\verb|7| & \Large \textcolor{red}{\xmark} & \Large \textcolor{red}{\xmark} \\
\hline
\end{tabular}
\end{table}

\begin{table}[t]
\centering
\caption{Example questions and evaluation comparison between \emphtxt{Lighteval framework} \cite{lighteval} and our LLM-as-a-judge approach  for the Qwen2.5-7B model. We observe that the symbolic evaluation limitations also exist in the popular Lighteval framework, while our approach generalizes to various answer representations. The questions are from the GSM8K dataset.
}
\label{tab:qexample3}
\setlength{\tabcolsep}{4pt}
\begin{tabular}{p{3cm} p{1.5cm} p{1.9cm}}
\hline
\multicolumn{3}{p{7.2cm}}{\small \textbf{Question:} "Becky bought 20 apples for 45 cents each and received a \$1 discount. Kelly bought 20 apples for 50 cents each and received a 10 percent discount. How much more did Kelly pay than Becky?" \; \textcolor{navyblue}{GT: 1} \; {\footnotesize (GSM8K)}} \\
\hline
Answer \small (Lighteval) & Symbolic & LLM \small (Ours) \\
\hline
\formatcorrecttext & & \\
\small\verb|1| & \Large \textcolor{darkgreen}{\cmark} & \Large \textcolor{darkgreen}{\cmark} \\
\small\verb|100| & \Large \textcolor{red}{\xmark} & \Large \textcolor{darkgreen}{\cmark} \\
\small\verb|1.00| & \Large \textcolor{darkgreen}{\cmark} & \Large \textcolor{darkgreen}{\cmark} \\
\formatincorrecttext & & \\
\small\verb|0.50| & \Large \textcolor{red}{\xmark} & \Large \textcolor{red}{\xmark} \\
\small\verb|0.30| & \Large \textcolor{red}{\xmark} & \Large \textcolor{red}{\xmark} \\
\small\verb|1.90| & \Large \textcolor{red}{\xmark} & \Large \textcolor{red}{\xmark} \\
\hline\hline
\multicolumn{3}{p{7.2cm}}{\small \textbf{Question:} "A family of 6 (2 adults and 4 kids) are to divide a watermelon such that each adult gets a slice that is twice as big as that of each kid. What percentage of the watermelon does each adult get?" \; \textcolor{navyblue}{GT: 25} \; {\footnotesize (GSM8K)}} \\
\hline
Answer \small (Lighteval) & Symbolic & LLM \small (Ours) \\
\hline
\formatcorrecttext & & \\
\small\verb|25%| & \Large \textcolor{red}{\xmark} & \Large \textcolor{darkgreen}{\cmark} \\
\small\verb|25| & \Large \textcolor{darkgreen}{\cmark} & \Large \textcolor{darkgreen}{\cmark} \\
\small\verb|25.0| & \Large \textcolor{red}{\xmark} & \Large \textcolor{darkgreen}{\cmark} \\
\formatincorrecttext & & \\
\small\verb|20| & \Large \textcolor{red}{\xmark} & \Large \textcolor{red}{\xmark} \\
\hline\hline
\multicolumn{3}{p{7.2cm}}{\small \textbf{Question:} "Stephen placed an online order for groceries. His final bill came to \$40.00. Because this was through a delivery vendor, they tacked on a 25\% fee to his final total and charged him \$3.00 in delivery fees. Stephen also added a \$4.00 tip. After the extra fees, what was the final price of Stephen's groceries?" \; \textcolor{navyblue}{GT: 57} \; {\footnotesize (GSM8K)}} \\
\hline
Answer & Symbolic & LLM \small (Ours) \\
\hline
\formatcorrecttext & & \\
\small\verb|57| & \Large \textcolor{darkgreen}{\cmark} & \Large \textcolor{darkgreen}{\cmark} \\
\small\verb|57.00| & \Large \textcolor{red}{\xmark} & \Large \textcolor{darkgreen}{\cmark} \\
\small\verb|$57.00| & \Large \textcolor{red}{\xmark} & \Large \textcolor{darkgreen}{\cmark} \\
\formatincorrecttext & & \\
\small\verb|54| & \Large \textcolor{red}{\xmark} & \Large \textcolor{red}{\xmark} \\
\small\verb|54.00| & \Large \textcolor{red}{\xmark} & \Large \textcolor{red}{\xmark} \\
\small\verb|58.75| & \Large \textcolor{red}{\xmark} & \Large \textcolor{red}{\xmark} \\
\small\verb|47| & \Large \textcolor{red}{\xmark} & \Large \textcolor{red}{\xmark} \\
\small\verb|47.00| & \Large \textcolor{red}{\xmark} & \Large \textcolor{red}{\xmark} \\
\small\verb|53| & \Large \textcolor{red}{\xmark} & \Large \textcolor{red}{\xmark} \\
\small\verb|57.75| & \Large \textcolor{red}{\xmark} & \Large \textcolor{red}{\xmark} \\
\hline
\end{tabular}
\end{table}

\begin{table}[t]
\centering
\caption{Example questions and evaluation comparison between \emphtxt{Lighteval framework} \cite{lighteval} and our LLM-as-a-judge approach for the \emphtxt{Llama3.1-8B model}. We observe that the symbolic evaluation limitations also exist in the popular Lighteval framework, while our approach generalizes to various answer representations. The questions are from the Math500 dataset.}
\label{tab:qexample4}
\setlength{\tabcolsep}{4pt}
\begin{tabular}{p{3cm} p{1.5cm} p{1.9cm}}
\hline
\multicolumn{3}{p{7.2cm}}{\small \textbf{Question:} "Write $\frac{3}{20}$ as a decimal. Answer: Let's think step by step." \; \textcolor{navyblue}{GT: 0.15} \; {\footnotesize (Math500)}} \\
\hline
Answer \small (Lighteval) & Symbolic & LLM \small (Ours) \\
\hline
\formatcorrecttext & & \\
\small\verb|\frac{3}{20} = 0.15| & \Large \textcolor{red}{\xmark} & \Large \textcolor{darkgreen}{\cmark} \\
\small\verb|0.15| & \Large \textcolor{darkgreen}{\cmark} & \Large \textcolor{darkgreen}{\cmark} \\
\hline\hline
\multicolumn{3}{p{7.2cm}}{\small \textbf{Question:} "What is the value of $(3x-2)(4x+1)-(3x-2)4x+1$ when $x=4$? Answer: Let's think step by step." \; \textcolor{navyblue}{GT: 11} \; {\footnotesize (Math500)}} \\
\hline
Answer \small (Lighteval) & Symbolic & LLM \small (Ours) \\
\hline
\formatcorrecttext & & \\
\small\verb|-160 + 1 + 170 = 11| & \Large \textcolor{red}{\xmark} & \Large \textcolor{darkgreen}{\cmark} \\
\small\verb|11| & \Large \textcolor{darkgreen}{\cmark} & \Large \textcolor{darkgreen}{\cmark} \\
\hline\hline
\multicolumn{3}{p{7.2cm}}{\small \textbf{Question:} "Let $S$ be the union of the set of all points inside a regular nonagon with side length 2 units and the set of all points less than 1 unit away from a point on the perimeter of the nonagon. What, in units, is the perimeter of $S$? Answer: Let's think step by step." \; \textcolor{navyblue}{GT: $2\pi + 18$}  \; {\footnotesize (Math500)}} \\
\hline
Answer \small (Lighteval) & Symbolic & LLM \small (Ours) \\
\hline
\formatcorrecttext & & \\
\small\verb|24.28| & \Large \textcolor{red}{\xmark} & \Large \textcolor{darkgreen}{\cmark} \\
\small\verb|24.283| & \Large \textcolor{red}{\xmark} & \Large \textcolor{darkgreen}{\cmark} \\
\formatincorrecttext & & \\
\small\verb|74.699| & \Large \textcolor{red}{\xmark} & \Large \textcolor{red}{\xmark} \\
\small\verb|18| & \Large \textcolor{red}{\xmark} & \Large \textcolor{red}{\xmark} \\
\small\verb|12| & \Large \textcolor{red}{\xmark} & \Large \textcolor{red}{\xmark} \\
\small\verb|29.773| & \Large \textcolor{red}{\xmark} & \Large \textcolor{red}{\xmark} \\
\small\verb|30.56| & \Large \textcolor{red}{\xmark} & \Large \textcolor{red}{\xmark} \\
\small\verb|74.52| & \Large \textcolor{red}{\xmark} & \Large \textcolor{red}{\xmark} \\
\hline
\end{tabular}
\end{table}

\begin{table}[t]
\centering
\caption{
Example cases where the \emphtxt{reference answers may be inaccurate due to ambiguous problem statements}. These examples demonstrate the challenge of evaluation when dataset GT answers are questionable and highlight cases where dataset answer validation is required. The questions are from the GSM8K dataset.}
\label{tab:ambiguous_example}
\setlength{\tabcolsep}{4pt}
\begin{tabular}{p{3cm} p{1.5cm} p{1.9cm}}
\hline
\multicolumn{3}{p{7.2cm}}{\small \textbf{Question:} "John orders some pizzas to share with his friends. There are 20 friends in total, and John wants to make sure each can have 4 slices. Pizzas are only sold sliced into 8 portions. How many Pizzas does John need to order?" \; \textcolor{navyblue}{GT: 10} \; {\footnotesize (GSM8K)}} \\
\hline
Answer & Symbolic & LLM \small (Ours) \\
\hline
\small\verb|10| & \Large \textcolor{darkgreen}{\cmark} & \Large \textcolor{red}{\xmark} \\
\small\verb|11| & \Large \textcolor{red}{\xmark} & \Large \textcolor{darkgreen}{\cmark} \\
\small\verb|80| & \Large \textcolor{red}{\xmark} & \Large \textcolor{red}{\xmark} \\
& & \\
\multicolumn{3}{p{7.2cm}}{\small{\emphtxt{Explanation:}} \footnotesize {The ambiguity arises from whether "each can have 4 slices" refers to the 20 friends only (requiring 10 pizzas) or includes John himself (21 people total, requiring 11 pizzas). The question wording is unclear about John's inclusion in the count.}} \\
\hline\hline
\multicolumn{3}{p{7.2cm}}{\small \textbf{Question:} "Steve decides to start eating more tomatoes and decides to grows his own cherry tomatoes. He eats twice as much as his girlfriend. He eats 6 per day. If a vine can produce 3 tomatoes per week how many vines does he need?" \; \textcolor{navyblue}{GT: 21} \; {\footnotesize (GSM8K)}} \\
\hline
Answer & Symbolic & LLM \small (Ours) \\
\hline
\small\verb|21| & \Large \textcolor{darkgreen}{\cmark} & \Large \textcolor{red}{\xmark} \\
\small\verb|14| & \Large \textcolor{red}{\xmark} & \Large \textcolor{darkgreen}{\cmark} \\
\small\verb|3| & \Large \textcolor{red}{\xmark} & \Large \textcolor{red}{\xmark} \\
\small\verb|42| & \Large \textcolor{red}{\xmark} & \Large \textcolor{red}{\xmark} \\
\small\verb|2| & \Large \textcolor{red}{\xmark} & \Large \textcolor{red}{\xmark} \\
\small\verb|19.37| & \Large \textcolor{red}{\xmark} & \Large \textcolor{red}{\xmark} \\
\multicolumn{3}{p{7.2cm}}{\small\emphtxt{Explanation:} \footnotesize {The question asks "how many vines does he need" specifically for Steve's consumption only (6 tomatoes/day = 42/week, requiring 14 vines), not for both Steve and his girlfriend combined (which would require 21 vines). The reference answer incorrectly includes the girlfriend's consumption.}} \\
\hline
\end{tabular}
\end{table}

\end{document}